\documentclass[runningheads]{llncs}

\usepackage{eccv}

\usepackage{eccvabbrv}

\usepackage{graphicx}
\usepackage{threeparttable}
\usepackage{caption}
\usepackage{booktabs}

\usepackage[accsupp]{axessibility}  

\usepackage[pagebackref,breaklinks,colorlinks,citecolor=eccvblue]{hyperref}

\usepackage{orcidlink}
\usepackage{amsfonts}       
\usepackage{nicefrac}       
\usepackage{microtype}      
\usepackage[dvipsnames]{xcolor}         
\usepackage{caption}
\usepackage{subcaption}
\usepackage{multirow}
\usepackage{colortbl}
\usepackage{soul}
\usepackage[ruled,vlined,linesnumbered]{algorithm2e}
\usepackage{fontawesome5}

\begin{document}

\title{Combating Textual Noise and Redundancy: Entropy-Aware Dense Visual Token Pruning} 

\titlerunning{Entropy-Aware Dense Visual Token Pruning}

\author{Xuehui Wang\inst{}$^{\dagger}$ \orcidlink{0000-0002-6333-7773} \and
Xuankun Yang\inst{}$^{\dagger}$ \orcidlink{0009-0001-0763-5776} \and
Wei Shen\inst{}$^\text{\faEnvelope}$ \orcidlink{0000-0002-1235-598X}}

\def\customsymbol#1{
    \ifcase\number\value{#1}
        \or*
        \or\faEnvelope
    \else\@ctrerr
    \fi
}

\authorrunning{X.~Wang et al.}

\institute{Shanghai Jiao Tong University, Shanghai, China \\
\email{\{wangxuehui,kk-dao,wei.shen\}@sjtu.edu.cn}\\
Codes: \url{https://github.com/SJTU-DeepVisionLab/EADP}
}

\maketitle
\renewcommand{\footnotesize}{\fontsize{8pt}{8pt}\selectfont}
\renewcommand{\thefootnote}{\customsymbol{footnote}}
\footnotetext[0]{${\dagger}$~Equal contribution. ~~\faEnvelope ~Corresponding author.}

\begin{abstract}
Visual token pruning is a crucial strategy for accelerating VLMs by compressing redundant image patches, yet existing methods often fail to preserve critical cues under dense instructions and fine-grained queries. In this paper, we investigate this failure and identify two underlying bottlenecks: the widespread dispersion of textual noise that corrupts dense cross-modal scoring, and the feature fragmentation inherent to standard token selection. To address these issues, we propose \textbf{E}ntropy-\textbf{A}ware \textbf{D}ense \textbf{P}runing (EADP), a framework that reformulates pruning as a structured compression problem. EADP first leverages statistical entropy to quantify and filter out textual noise, yielding a robust, fine-grained instruction relevance score. Subsequently, instead of naive Top-$K$ selection, EADP casts token selection as a submodular maximization problem with a spatial prior, explicitly ensuring a holistic and non-redundant visual representation. Extensive experiments demonstrate that EADP improves the accuracy-efficiency trade-off of VLMs, robustly preserving fine-grained visual cues under strict token budgets while achieving SoTA performance on challenging multimodal benchmarks.

\end{abstract}

\section{Introduction}
\label{sec:intro}

Vision-language models (VLMs)~\cite{liu2023llava,liu2024llavanext,chen2024internvl2.5,bai2025qwen25vltechnicalreport,lin2024videollava,li2024llava-onevision,guan2025token,guan2026codepercept} are foundational for multimodal understanding, yet their deployment is hindered by the high computational cost of long visual contexts. In particular, modern VLMs represent images as dense grids of patch tokens, directly driving up quadratic self-attention costs and end-to-end latency~\cite{vaswani2017attention,dosovitskiy2021imageworth16x16words,zhang2024sparsevlm}. This makes \emph{visual token pruning} an appealing and increasingly necessary strategy for real-time and resource-constrained applications: retaining only a compact subset of informative tokens significantly accelerates inference with minimal performance degradation~\cite{shang2025prumerge,chen2024fastv,ye2025fitprune,song2025trim,duan2025gridprune,lin2025vtw,yang2025visionzip,alvar2025divprune,zhang2026pio,wen2025dart}. 

In practice, however, things rarely go that smoothly. When we push pruning to the regimes that actually matter for deployment, tight budgets, dense instructions, fine-grained cues, and challenging negative queries, existing pruning systems often become brittle, easily discarding small yet decisive visual cues. This raises a deceptively simple question: \emph{what exactly goes wrong when we prune?}
To answer this, we dissect the standard visual pruning pipeline, which typically consists of two stages: scoring token importance and selecting the Top-$K$ tokens~\cite{liu2025hiprune, zou2025dontjustchasehighlighted}. In the scoring stage, the prevailing recipe relies on a single global text feature (e.g., the CLIP~\cite{radford2021clip} \texttt{EOS} token) to evaluate each visual token~\cite{zhang2025cdpruner,zhang2025trimtokenator,li2025catp,zhang2025trimtokenatorlcadaptivevisualtoken}. While cheap and stable, it often acts as a highly compressed summary, missing fine-grained details. The intuitive fix for such coarse global guidance is dense guidance, computing cross-modal similarity between \emph{every} text token and visual token. Surprisingly, this ``more informative'' strategy often refuses to help. We observe that functional words and punctuation typically produce broadly dispersed, near-uniform similarity responses~\cite{clark-etal-2019-bert,vig-belinkov-2019-analyzing}. Together, they accumulate into an indiscriminate ``noise floor'' that drowns out the sparse, localized peaks of truly meaningful semantic entities. Moreover, even if we succeed in obtaining a clean, fine-grained relevance score, a second surprise awaits at the selection stage. The standard Top-$K$ rule, though universally used, is a flawed compression strategy under strict budgets. It tends to over-concentrate on a few highly discriminative regions, wasting the limited token budget on redundant local maxima (e.g., repeatedly sampling the head of a target object) while leaving other integral semantic parts entirely unrepresented.

These observations reveal that visual pruning is not merely a scoring task, but a structured compression problem requiring both noise-robust saliency and holistic visual coverage. 
To address this, we propose \textbf{Entropy-Aware Dense Pruning (EADP)}, a lightweight framework designed to make instruction relevance fine-grained and token selection non-redundant. 
Concretely, EADP first reformulates the instruction relevance scoring into a dual-stream process. We begin by computing dense cross-modal similarities between all non-\texttt{EOS} CLIP text tokens and VLM visual tokens. To eliminate the dispersed textual noise identified earlier, we introduce an entropy-guided denoising mechanism that directly calculates the entropy of each text token's spatial similarity distribution. By filtering out high-entropy tokens and weighting the retained low-entropy ones, we obtain a clean dense guidance score. Finally, we fuse this denoised dense signal with the global \texttt{EOS} semantic context, yielding a comprehensive instruction relevance map that possesses both local precision and macroscopic stability.

With a robust instruction relevance map in hand, EADP proceeds to tackle the feature fragmentation and redundancy caused by standard Top-$K$ selection. We first refine the score map to preserve spatial integrity: a lightweight Gaussian smoothing is applied to incorporate local structural context, followed by a score polarization step that exponentially re-amplifies the core visual entities against the smoothed background. Armed with these structurally aware scores, we abandon the naive Top-$K$ heuristic and cast the final token selection as a facility location submodular maximization problem~\cite{Nemhauser1978AnAO,Krause2014SubmodularFM,Lin2011ACO}. This mathematically principled objective shifts the focus from mere peak-chasing to holistic representativeness. It guarantees that all semantic parts of the original image are well-covered by the compressed subset, naturally penalizing redundant tokens and yielding a compact, highly informative visual representation for the downstream LLM.

In summary, our main contributions are three-fold:
\begin{itemize}
    \item Crucial Insights into Pruning Bottlenecks: We systematically analyze the failure modes of existing visual token pruning paradigms. We identify the dispersion phenomenon of textual noise in dense guidance, and reveal feature fragmentation and redundancy inherent to standard Top-$K$ selection.
    \item Entropy-Aware Dense Scoring: We propose a novel dispersion-aware scoring mechanism that elegantly quantifies and filters textual noise using statistical entropy. Without relying on external NLP parsers, this approach extracts highly precise, fine-grained instruction relevance and fuses it with global semantic context for robust guidance.
    \item Submodular Token Selection with Spatial Priors: We reformulate visual token selection as a facility location submodular maximization problem. Coupled with a spatial smoothing and score polarization prior, this objective explicitly guarantees holistic feature representativeness and penalizes local redundancy, fundamentally overcoming the limitations of greedy Top-$K$ sampling.
\end{itemize}

\section{Related Works}
\label{sec:related_works}

\begin{figure}[t]
  \centering
  \includegraphics[width=\linewidth]{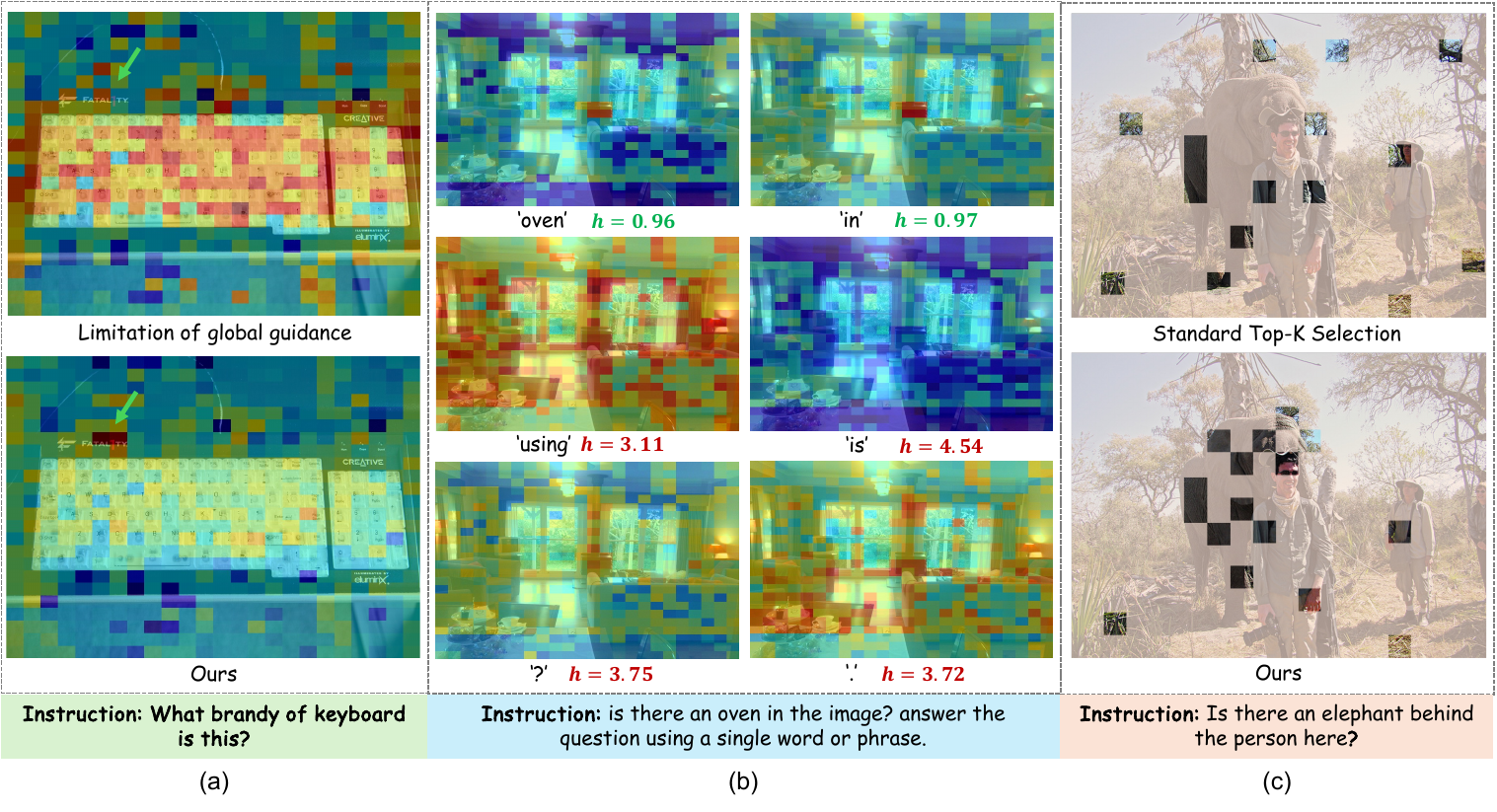}
  \caption{(a) illustrates a limitation of global guidance: it tends to attend to background regions. (b) highlights the dispersion phenomenon caused by textual noise. (c) reveals the issues of feature fragmentation and selection redundancy.
  }
  \label{fig:motivation}
\end{figure}

\textbf{Large Vision-Language Models.}
Recent Vision-Language Models (VLMs) extend LLMs with visual encoders to enable robust multimodal instruction following~\cite{radford2021clip,alayrac2022flamingo,li2023blip2}. 
Building upon early contrastive pretraining and unified architectures~\cite{radford2021clip,jia2021align,cho2014gru,vaswani2017attention,li2020oscar}, current paradigms predominantly align pretrained vision backbones with LLMs via lightweight connectors and multimodal instruction tuning~\cite{liu2023llava,zhu2023minigpt4,ye2023mplugowl,li2023instructblip}. 
Further scaling and advanced prompting techniques (e.g., Chain-of-Thought) have unlocked emergent capabilities in fine-grained and compositional reasoning~\cite{alayrac2022flamingo,bai2023qwenvl,chen2023internvl,wei2022cot,kojima2022zeroshotcot}. 
However, the pursuit of higher-resolution perception inevitably produces massive visual token sequences. The resulting quadratic computational overhead poses a severe bottleneck for practical inference and deployment~\cite{li2023blip2,liu2023llava}, urgently motivating research into compact representations and visual token pruning.

\noindent \textbf{Visual Token Pruning.}
Visual token pruning aims to drop redundant patches while preserving task-relevant information, broadly following three paths. 
\emph{(i) Score-/saliency-based pruning}~\cite{shang2025prumerge,chen2024fastv,ye2025fitprune,zhang2024sparsevlm,song2025trim,zhang2025cdpruner,liu2025hiprune} evaluates token importance to retain the top-$k$ subset, utilizing heuristics, cross-modal dependency, or lightweight predictors. 
\emph{(ii) Structure-aware pruning}~\cite{duan2025gridprune,lin2025vtw,yang2025visionzip,alvar2025divprune,zhang2026pio} exploits spatial layouts by grouping patches or selecting informative regions to maintain structural coverage and diversity. 
\emph{(iii) Training-/policy-based pruning}~\cite{wen2025dart,li2025catp,zhang2025trimtokenator} learns dynamic, context-adaptive token-dropping decisions end-to-end. 
Despite promising speedups, existing methods heavily rely on coarse global guidance and heuristic Top-K sampling, suffering from textual noise and feature fragmentation. In contrast, our EADP resolves these bottlenecks by filtering dispersed dense textual noise via Information Entropy and mathematically guaranteeing holistic feature representativeness through Submodular Maximization.

\section{Observations and Insights}
\label{sec:rethinking}

Recent methods~\cite{zhang2025cdpruner,duan2025gridprune,zhang2025trimtokenator} tend to combine feature similarity among visual tokens with instruction relevance to derive the importance score for each visual token and then sample a subset of high score tokens as the final visual tokens.
In this section, we conduct pilot studies to revisit existing SoTA pruning paradigms, revealing critical bottlenecks in both importance scoring and token selection.

\subsection{The Limitation of Global Guidance}
\label{sec:limitation_of_global}

We utilize CDPruner, which adopts the \texttt{EOS} token from the CLIP text encoder as the global guidance token to compute instruction relevance, to investigate the efficacy of global guidance.
In our experiment, we visualize the instruction relevance maps generated by CDPruner on dense visual tasks, as shown in~\cref{fig:motivation}(a).
We notice that the global guidance token tends to highlight broad, semantically vague background regions and does not consistently concentrate on fine-grained critical clues.
This observation suggests that the global guidance token acts as a highly compressed semantic summary, which inherently lacks the resolution required to capture fine-grained textual instructions~\cite{yao2021filipfinegrainedinteractivelanguageimage,li2022groundedlanguageimagepretraining}. 
Consequently, global-guided pruning can be sub-optimal for fine-grained recognition and challenging negative queries where the referenced object is absent from the image.

\begin{figure}[t]
  \centering
  \includegraphics[width=\linewidth]{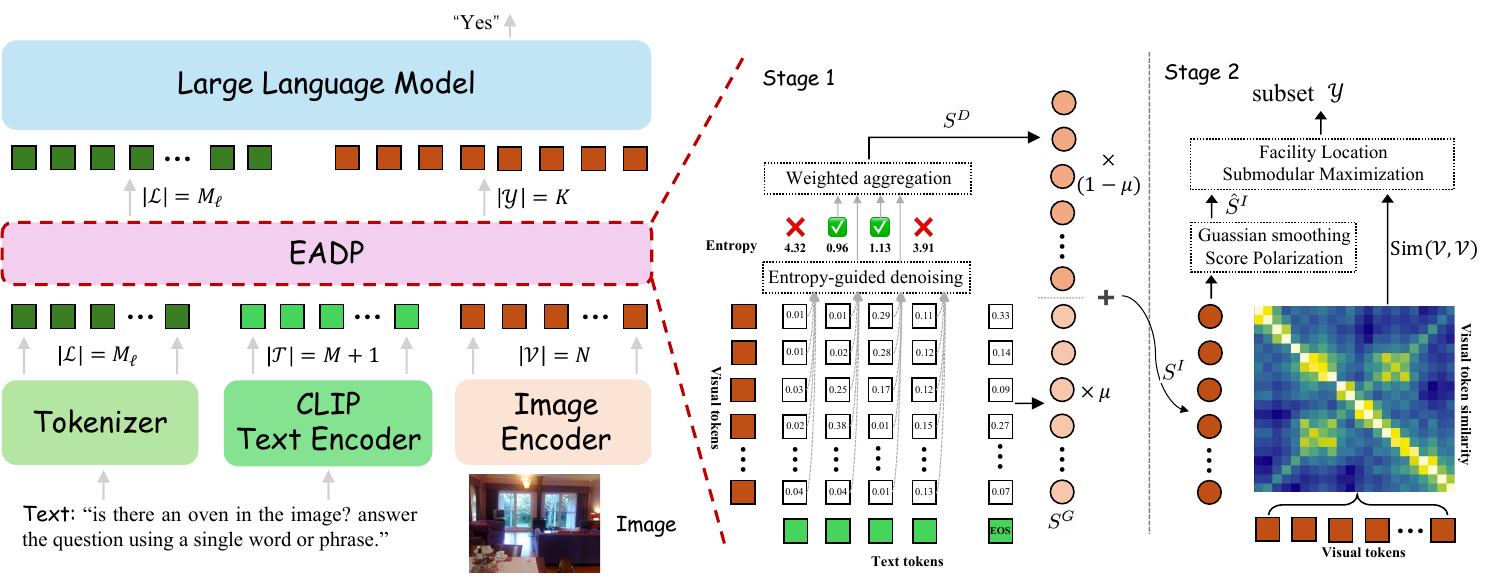}
  \vspace{-2mm}
  \caption{
  \textbf{Overview of the EADP.} EADP acts as a plug-and-play module compressing $N$ visual tokens into a highly informative subset of $K$ tokens for the downstream LLM. 
  \textbf{Stage 1}: An entropy-guided denoising mechanism filters out high-entropy textual noise to get the dense guidance score $S^D$. This is fused with the global \texttt{EOS} score $S^G$ to yield a robust instruction relevance score $S^I$. \textbf{Stage 2}: After refining $S^I$ via gaussian smoothing and score polarization to prevent feature fragmentation, a submodular maximization objective uses inter-token visual similarity to select the final subset $\mathcal{Y}$.
  }
  \vspace{-3mm}
  \label{fig:EADP}
\end{figure}

\subsection{The Dispersion Phenomenon of Textual Noise}
\label{sec:dispersion_phenomenon}

An intuitive solution to the aforementioned limitation is dense guidance: calculating the cross-modal similarity between every text token in the prompt and the visual tokens. 
However, our empirical results reveal a counter-intuitive phenomenon: naive dense aggregation yields no performance improvement.
To understand why dense guidance fails, we visualize the spatial similarity distribution induced by individual text tokens on visual tokens and we uncover a distinct Dispersion Phenomenon in cross-modal similarity: as illustrated in~\cref{fig:motivation}(b), semantic entity words (e.g., "oven", "cat") exhibit a highly concentrated similarity distribution, accurately localizing their corresponding visual patches. In stark contrast, functional words~\cite{clark-etal-2019-bert,rogers-etal-2020-primer,xiao2024efficientstreaminglanguagemodels} (e.g., "using", "is") and punctuation marks (e.g., "?") exhibit a uniformly dispersed similarity map across the entire image.

While an individual dispersed text token produces uniformly low response values across visual patches, their cumulative effect is devastating. Since functional words and punctuation typically constitute the majority of a text prompt, aggregating their similarities creates a dominating, indiscriminate ``noise floor’’. This textual noise inflates the scores of irrelevant background regions, diluting and submerging the localized activation peaks generated by the few semantic entity words. Identifying and filtering out these dispersed text tokens, without relying on NLP parsers, is a prerequisite for effective dense pruning. We later show that statistical entropy can quantify this dispersion degree in~\cref{sec:dense_scoring}.

\subsection{Feature Fragmentation and Selection Redundancy}
\label{sec:fragmentation_selection}

Even if we could perfectly filter the textual noise and obtain accurate importance scores, the final step, selecting a compact subset of visual tokens, remains non-trivial. To isolate the selection issue, we perform a controlled analysis: we manually choose the prompt words that refer to the target object and use their induced similarity scores as instruction relevance, combined with inter-token feature similarity, to derive the final importance scores. Subsequently, we apply the standard Top-K selection based on these scores.

As illustrated in~\cref{fig:motivation}(c), we observe a severe ``feature fragmentation'' issue. Rather than capturing the holistic structure of the referring object, the Top-K selection tends to over-fit to the most highly discriminative local regions (e.g., the tail of a elephant), leaving other integral parts (e.g., the body and head) entirely unrepresented. Furthermore, this naive selection inevitably leads to severe redundancy, where selected tokens heavily cluster within a few high-score peaks, wasting the limited token budget on repetitive local features while completely missing the broader semantic context.
These observations highlight three requirements for a robust pruning framework:
(1) a spatial smoothing mechanism to propagate activations and incorporate local context,
(2) a contrast sharpening operation to highlight core visual entities and suppress smoothed background noise,
and (3) a principled selection paradigm that guarantees holistic feature representativeness, ensuring all semantic parts are covered, while penalizing local redundancy.
This finding motivates our adoption of a gaussian spatial prior, score polarization, and submodular maximization in~\cref{sec:smoothing_selection}.

\vspace{-1mm}
\section{Methodology}
\label{sec:methodology}

In this section, we present our Entropy-Aware Dense Pruning (EADP) framework. Building upon the insights from~\cref{sec:rethinking}, EADP explicitly quantifies textual dispersion to filter out noisy text tokens, fuses dense and global guidance for robust instruction relevance scoring, and reformulates token selection as a submodular maximization problem to preserve spatial integrity and eliminate redundancy. The overview pipeline of EADP is shown in ~\cref{fig:EADP}.

\subsection{Preliminaries and Problem Formulation}
Given an input image $I$ and a task-specific text prompt $P$, a VLM typically employs a vision encoder to extract a sequence of visual tokens $\mathcal{V} = \{v_1, v_2, ..., v_N\} \in \mathbb{R}^{N \times d_v}$, where $N$ is the number of visual patches and $d_v$ is the feature dimension. 
Meanwhile, the LLM tokenizer converts the text prompt into language tokens for generation; we denote these LLM tokens as $\mathcal{L} = \{l_1, l_2, \dots, l_{M_\ell}\}$.
These LLM tokens are concatenated with projected visual tokens and fed into the LLM to autoregressively generate the output.

We additionally encode the same prompt $P$ by a CLIP text encoder to obtain a sequence of CLIP text features
$\mathcal{T} = \{t_1, t_2, \dots, t_{M}, t_{\texttt{EOS}}\} \in \mathbb{R}^{(M+1)\times d}$,
where $t_{\texttt{EOS}}$ is the \texttt{EOS} token feature and $d$ is the CLIP embedding dimension.
To compute cross-modal cosine similarities between $\mathcal{T}$ and visual tokens from the VLM vision tower, we project visual tokens to the CLIP embedding space via a lightweight projector (implemented identically to CDPruner), producing $\tilde{\mathcal{V}}=\{\tilde{v}_j\}_{j=1}^N$ with $\tilde{v}_j\in\mathbb{R}^d$.
For simplicity, we use $v_j$ to denote the projected visual token when computing cross-modal similarities below.

The goal of visual token pruning is to identify a compact subset $\mathcal{Y} \subset \mathcal{V}$ with size $|\mathcal{Y}| = K \ll N$ that preserves the most informative visual cues required for correct generation.
This requires (i) an instruction relevance scoring function $\mathcal{F}: v_j \mapsto \mathbb{R}$ and (ii) a selection strategy $\Phi$:
\begin{equation}
    \mathcal{Y} = \Phi(\mathcal{F}(\mathcal{V}); K) \in \mathbb{R}^{K \times d_v}.
\end{equation}
We detail the formulation of $\mathcal{F}$ and $\Phi$ in~\cref{sec:dense_scoring} and~\cref{sec:smoothing_selection}, respectively.

\subsection{Dispersion-Aware Dense Scoring and Fusion}
\label{sec:dense_scoring}

In EADP, instruction relevance is computed from two complementary components: a global guidance score following prior work~\cite{zhang2025cdpruner,duan2025gridprune}, and our proposed dispersion-aware dense guidance score.
Both are derived from cross-modal cosine similarities between CLIP text features $\mathcal{T}$ and projected visual tokens $\mathcal{V}$.

\noindent \textbf{Global Guidance Score.} We compute similarities between the \texttt{EOS} token and visual tokens as in CDPruner~\cite{zhang2025cdpruner}, yielding a global guidance score $S^G \in \mathbb{R}^N$:
\begin{equation}
    s_j^G = \frac{v_j \cdot t_{\texttt{EOS}}}{\lVert v_j\rVert \cdot \lVert t_{\texttt{EOS}}\rVert}, \quad v_j \in \mathcal{V}.
\end{equation}

\noindent \textbf{Dispersion-aware Dense Guidance Score.} As observed in~\cref{sec:limitation_of_global,sec:dispersion_phenomenon}, relying solely on the global token $t_{EOS}$ neglects fine-grained details, while naive dense guidance introduces severe textual dispersion noise from functional words. To address this, we introduce an entropy-guided denoising mechanism. We first compute the dense cosine similarity matrix $C \in \mathbb{R}^{M \times N}$ between \emph{non-\texttt{EOS}} CLIP text tokens $\{t_i\}_{i=1}^{M}$ and all visual tokens $\{v_j\}_{j=1}^{N}$:
\begin{equation}
    c_{i,j} = \frac{t_i \cdot v_j}{\lVert t_i\rVert \cdot \lVert v_j\rVert}, \quad i\in\{1,\dots,M\},\; j\in\{1,\dots,N\}.
\end{equation}
Since $M \le 77$ in CLIP, calculating $C$ requires just one lightweight matrix multiplication, adding negligible overhead compared to the LLM attention.

To measure the dispersion degree of a text token's attention, we convert its similarity scores over all visual tokens into a probability distribution via a temperature-controlled Softmax:
\begin{equation}
    p_{i,j} = \frac{\exp(c_{i,j} / \tau)}{\sum_{k=1}^N \exp(c_{i,k} / \tau)}.
\end{equation}
We then calculate the Information Entropy $H \in \mathbb{R}^{M}$ for the tokens:
\begin{equation}
    h_i = - \sum_{j=1}^N p_{i,j} \log p_{i,j}, \ \ H = \{h_1, h_2, ..., h_M\}
\end{equation}
A higher entropy $h_i$ indicates that the attention of $t_i$ is uniformly dispersed across the image (i.e., textual noise, such as punctuation or functional words), whereas a lower $h_i$ signifies a highly concentrated attention on specific visual entities. 

We retain a proportion $q\in(0,1]$ of text tokens with the \emph{lowest} entropy.
Specifically, let $\lambda = Q_q(H)$ denote the $q$-quantile (lower tail) of entropy values; we keep tokens satisfying $h_i\le \lambda$:
\begin{equation}
    \mathcal{T}' = \{t_i \in \mathcal{T} | h_i \leq \lambda\}, \quad \lambda=Q_{q}(H),
\end{equation}
Next, we assign each retained text token in $\mathcal{T}'$ an entropy-based weight:
\begin{equation}
    \alpha_i = \texttt{softmax}_{i\in\mathcal{T}'}(-h_i / \gamma)
    \label{eq:aggregation_method}
\end{equation}
where $\gamma>0$ controls how strongly low-entropy tokens are emphasized.
The dispersion-aware dense guidance score $S^{D}\in\mathbb{R}^{N}$ is computed as a weighted aggregation of similarities:
\begin{equation}
    s_{j}^D = \sum_{i \in \mathcal{T}'} \alpha_i \cdot c_{i,j} 
\end{equation}

\noindent \textbf{Instruction Relevance Score.} While the dense guidance score $S^D$ excels at capturing fine-grained details, the global guidance score $S^G$ still provides a valuable macroscopic semantic context. 
Therefore, we fuse them to obtain the instruction relevance score $S^{I}\in\mathbb{R}^{N}$:
\begin{equation}
    s_j^I = \mu s_j^G + (1-\mu)s_j^D
\end{equation}
where $\mu \in [0, 1]$ is a balancing hyper-parameter. Furthermore, the instruction relevance score $S^I$ is applied to $\texttt{min-max}$ normalization to ensure that it lies in $[0,1]$, and then served as the foundational metric for our subsequent spatial smoothing and submodular selection stages:
\begin{equation}
    s_j^{I} = \frac{s_j^{I} - \min_{k\in\{1,\dots,N\}} s_k^{I}}
    {\max_{k\in\{1,\dots,N\}} s_k^{I} - \min_{k\in\{1,\dots,N\}} s_k^{I} + \varepsilon},
    \quad S^{I}=\{s_1^{I},\dots,s_N^{I}\},
\end{equation}
where $\varepsilon$ is a small constant for numerical stability.

\subsection{Spatial Smoothing Prior and Submodular Token Selection}
\label{sec:smoothing_selection}

To elegantly resolve the issues of feature fragmentation and selection redundancy stated in~\cref{sec:fragmentation_selection}, we introduce a spatial smoothing prior coupled with score polarization, followed by a submodular maximization sampling strategy.

\noindent \textbf{Spatial Continuity and Score Polarization.}
The fused relevance score $S^{I}$ is computed token-wise. To explicitly inject a spatial continuity prior, we reshape the 1D score vector $S^I\in\mathbb{R}^{N}$ back into its corresponding 2D spatial map $S^I_{\text{2D}} \in \mathbb{R}^{H \times W}$, where $N = H \times W$. Subsequently, we apply a lightweight $3 \times 3$ Gaussian convolutional kernel $G$ to smooth the score map, aggregating local structural context:
\begin{equation}
    S_{\text{smooth}}^I = S^I_{\text{2D}} * G.
\end{equation}
While spatial smoothing effectively propagates high activations to adjacent structural parts (e.g., providing local context around a highly discriminative region), it inherently acts as a low-pass filter that diminishes the peak scores of those discriminative tokens. To prevent critical features from being submerged, we introduce a non-linear score polarization step. We flatten $S^{I}_{\text{smooth}}$ back to a 1D vector $\{s^{I}_{\text{smooth},j}\}_{j=1}^{N}$ and apply an exponential sharpening filter:
\begin{equation}
    \hat{s}_j^{I} = \left(s^{I}_{\text{smooth},j}\right)^{\beta}, \quad \forall j\in\{1,\dots,N\},
\end{equation}
where $\beta > 1$ is a polarization factor. This operation exponentially amplifies the most salient regions while suppressing the smoothed background noise, yielding the final robust instruction relevance score $\hat{S}^I$.

\noindent \textbf{Facility Location Submodular Maximization.}
Given $\hat{S}^I$, our final objective is to sample a compact, non-redundant subset $\mathcal{Y} \subset \mathcal{V}$. To fundamentally align token selection with the essence of visual compression, i.e., representing the holistic with a few tokens, we reformulate the process as a classic Facility Location Submodular Maximization problem:

\begin{equation}
    \arg\max_{\mathcal{Y}\subset\mathcal{V}} \sum_{j=1}^{N} \hat{s}^I_j \cdot \max_{v_i\in\mathcal{Y}} \text{Sim}(v_i, v_j),
\end{equation}
where $\text{Sim}(v_i, v_j)$ denotes the cosine similarity between the visual features of token $v_i$ and $v_j$.
Unlike determinantal point processes (DPP)~\cite{macchi1975dpp}, which are used in CDPruner and primarily model diversity, the inner maximization term explicitly quantifies representativeness. It encourages that every single token $v_j$ in the original image is well-represented by its most similar counterpart in the selected subset $\mathcal{Y}$. This explicitly prevents over-fitting to localized high-score regions (e.g., redundantly sampling only the head of a target object) by ensuring holistic feature coverage across all semantic parts. Weighted by the highly polarized instruction relevance score $\hat{S}^I$, this objective forces the selection to prioritize salient regions, while the submodular maximization naturally penalizes selecting redundant tokens within those localized high-score areas.

While finding the exact optimal subset is NP-Hard, the facility location objective is inherently submodular, allowing us to employ a highly efficient greedy algorithm with a theoretical $1 - 1/e$ lower bound approximation. At each step, we iteratively add the candidate token $u \in \mathcal{V} \setminus \mathcal{Y}$ that yields the maximum marginal gain, computed as:

\begin{equation}
    \text{Gain}(u) = \sum_{j=1}^{N} \hat{s}^I_j \times (\max\left(\text{Sim}(u, v_j), \text{Curr}(v_j)) - \text{Curr}(v_j)\right)
\end{equation}
where $\text{Curr}(v_j) = \max_{v_i\in\mathcal{Y}} \text{Sim}(v_i, v_j)$ caches the maximum similarity between $v_j$ and the currently selected subset $\mathcal{Y}$. This caching allows efficient incremental updates of marginal gains.
In practice, the runtime depends on whether we precompute token-to-token similarities.
Computing cosine similarities on-the-fly yields a cost dominated by similarity evaluations, while optional precomputation of an $N\times N$ similarity matrix can accelerate greedy updates at the expense of $\mathcal{O}(N^2)$ memory.
\emph{We provide proofs and implementation details in the supplementary.}

\section{Experiments}
\label{sec:experiments}

\subsection{Experimental setup}

\noindent \textbf{Model.}
To comprehensively validate the effectiveness and generalizability of our method, we conduct experiments on multiple representative VLMs. Specifically, we adopt several variants from the LLaVA family, including LLaVA-1.5~\cite{liu2023llava} for standard image understanding, LLaVA-NEXT~\cite{liu2024llavanext} designed to handle high-resolution visual inputs, and LLaVA-Video~\cite{zhang2025llavavideovideoinstructiontuning} for video-based reasoning. In addition, we further evaluate EADP on Qwen2.5-VL~\cite{bai2025qwen25vltechnicalreport}, an advanced VLM. 
Results on more architectures are available in supplementary material.

\begin{table}[t]
  \centering
  \caption{\textbf{Performance comparison on LLaVA-1.5-7B.} Avg. denotes the average over 9 benchmarks (excluding VizWiz). * indicates results reproduced on the VizWiz validation set, since the official test challenge is no longer available.}
  \label{tab:llava-1.5}
  \resizebox{\linewidth}{!}{
    \begin{tabular}{l|cccccccccc|cc}
    \toprule
    \textbf{Method} & \textbf{$\text{VQA}^\text{V2}$} & \textbf{GQA} & \textbf{VizWiz} & \textbf{$\text{SQA}^\text{IMG}$} & \textbf{$\text{VQA}^\text{Text}$} & \textbf{POPE} & \textbf{MME} & \textbf{$\text{MMB}^\text{EN}$} & \textbf{$\text{MMB}^\text{CN}$} & \textbf{MMVet} & \textbf{Avg.} \\
    \rowcolor{lightgray!75}\multicolumn{12}{c}{\textit{Upper Bound, All 576 Tokens} ($\mathbf{100\%}$)} \\
    \rowcolor{lightgray!25} LLaVA-1.5-7B & 78.5 & 61.9 & 50.1/55.6* & 69.6 & 58.2 & 85.9 & 1513.4 & 64.7 & 58.1 & 31.3 & 64.9 \\
    \rowcolor{lightgray!75}\multicolumn{12}{c}{\textit{Retain 128 Tokens} \textcolor{Green}{($\downarrow\mathbf{77.8\%}$)}} \\
    FastV{\small\texttt{(ECCV24)}} & 71.3 & 55.3 & 51.9 & 68.7 & 55.6 & 70.0 & 1343.6 & 62.5 & 54.8 & 26.5 & 59.1 \\
    PDrop{\small\texttt{(CVPR25)}} & 73.7 & 56.8 & 49.4 & 69.5 & 56.1 & 77.9 & 1421.9 & 62.1 & 55.0 & 27.1 &  61.0 \\
    SparseVLM{\small\texttt{(ICML25)}} & 74.9 & 57.0 & 49.7 & 69.3 & 56.0 & 83.3 & 1408.3 & 62.2 & 56.1 & 30.1 & 62.1  \\
    PruMerge+{\small\texttt{(2024.05)}} & 74.8 & 58.0 & 53.7 & 68.6 & 54.3 & 83.2 & 1411.9 & 62.0 & 55.1 & 29.9 & 61.8 \\
    TRIM{\small\texttt{(COLING25)}} & 75.5 & 58.7 & 51.6 & 68.3 & 52.2 & 85.5 & 1418.5 & 61.9 & 52.9 & 29.5 & 61.7  \\
    VisionZip{\small\texttt{(CVPR25)}} & 75.5 & 57.9 & 51.6 & 68.9 & 55.9 & 84.8 & 1423.7 & 61.5 & 55.9 & 31.3 &  62.5 \\
    DART{\small\texttt{(EMNLP25)}} & 74.9 & 58.3 & 52.8 & 69.1 & 55.8 & 81.2 & 1415.1 & 60.7 & 56.0 & 31.1 & 62.0  \\
    DivPrune{\small\texttt{(CVPR25)}} & 76.0 & 59.2 & {57.5}$^*$ & 68.2 & 55.3 & 86.8 & 1405.1 & 61.5 & 54.8 & 30.6 & 62.5  \\
    CDPruner{\small\texttt{(NIPS25)}} & 76.2 & 59.3 & {57.3}$^*$ & 69.0 & 56.0 & 87.3 & 1431.4 & 62.4 & 54.7 & 32.1 & 63.2  \\
    HiPrune{\small\texttt{(AAAI26)}} & 74.9 & 57.3 & {55.4}$^*$ & 68.3 & 56.2 & 82.8 & 1364.4 & 62.2 & 56.4 & 31.2 & 61.9 \\
    \textbf{EADP}{\small\texttt{(Ours)}} & \textbf{76.7} & \textbf{60.0} & \textbf{58.0}$^*$ & \textbf{69.0} & \textbf{56.5} & \textbf{87.2} & \textbf{1439.0} & \textbf{62.7} & \textbf{55.2} & \textbf{32.5} & \textbf{63.5} \\
    \rowcolor{lightgray!75}\multicolumn{12}{c}{\textit{Retain 64 Tokens} \textcolor{Green}{($\downarrow\mathbf{88.9\%}$)}} \\
    FastV{\small\texttt{(ECCV24)}} & 57.2 & 48.0 & 49.1 & 68.7 & 52.1 & 37.2 &  992.1 & 49.6 & 43.3 & 20.1 & 47.3\\
    PDrop{\small\texttt{(CVPR25)}} & 58.1 & 49.2 & 46.3 & 68.3 & 51.1 & 40.1 &  1004.4 & 48.3 & 37.8 & 19.7 & 47.0  \\
    SparseVLM{\small\texttt{(ICML25)}} & 67.3 & 51.5 & 49.4 & 69.0 & 52.5 & 70.2 & 1192.9 & 58.0 & 49.3 & 25.1 & 55.8 \\
    PruMerge+{\small\texttt{(2024.05)}} & 71.2 & 55.1 & 53.7 & 69.2 & 52.1 & 76.3 & 1310.5 & 59.2 & 51.6 & 28.0 & 58.7  \\
    TRIM{\small\texttt{(COLING25)}} & 72.8 & 56.8 & 51.1 & 69.1 & 50.8 & 85.3 & 1356.1 & 60.5 & 49.7 & 26.5 & 59.9 \\
    VisionZip{\small\texttt{(CVPR25)}} & 72.1 & 56.0 & 52.9 & 69.0 & 55.0 & 77.9 & 1362.2 & 60.3 & 55.0 & 29.0 &  60.3 \\
    DART{\small\texttt{(EMNLP25)}} & 71.9 & 55.2 & 53.5 & 68.7 & 54.3 & 74.3 & 1371.3 & 59.3 & 54.0 & 26.5 & 59.2  \\
    DivPrune{\small\texttt{(CVPR25)}} & 74.1 & 57.5 & {57.8}$^*$ & 68.0 & 54.5 & 85.5 & 1356.2 & 60.1 & 52.3 & 28.1 & 60.9  \\
    CDPruner{\small\texttt{(NIPS25)}} & 75.0 & 58.6 & {58.0}$^*$ & 68.1 & 54.7 & 87.0 & 1399.1 & 61.1 & 53.2 & 29.5 & 61.9 \\
    HiPrune{\small\texttt{(AAAI26)}} & 69.2 & 53.6 & {55.4}$^*$ & 68.9 & 54.2 & 73.0 & 1236.0 & 59.5 & 53.4 & 27.9 & 57.9 \\
    \textbf{EADP}{\small\texttt{(Ours)}} & \textbf{75.6} & \textbf{59.4} & \textbf{59.5}$^*$ & \textbf{68.9} & \textbf{55.0} & \textbf{86.5} & \textbf{1403.6} & \textbf{61.9} & \textbf{53.1} & \textbf{29.4} & \textbf{62.2}\\
    \rowcolor{lightgray!75}\multicolumn{12}{c}{\textit{Retain 32 Tokens} \textcolor{Green}{($\downarrow\mathbf{94.4\%}$)}} \\
    PruMerge+{\small\texttt{(2024.05)}} & 65.3 & 51.3 & 53.5 & 68.1 & 48.7 & 68.9 & 1255.2 & 54.6 & 45.2 & 25.5 & 54.5  \\
    TRIM{\small\texttt{(COLING25)}} & 68.9 & 53.7 & 50.7 & 68.0 & 46.5 & 84.2 & 1259.3 & 57.1 & 39.5 & 22.8 &  56.0 \\
    VisionZip{\small\texttt{(CVPR25)}} & 67.4 & 52.2 & 52.4 & 68.7 & 52.0 & 70.6 & 1257.1 & 56.8 & 48.9 & 24.8 & 56.0 \\
    DART{\small\texttt{(EMNLP25)}} & 67.3 & 52.6 & 52.5 & 69.1 & 52.2 & 71.1 & 1288.4 & 58.5 & 49.3 & 25.5 & 56.7 \\
    DivPrune{\small\texttt{(CVPR25)}} & 71.2 & 54.9 & {57.4}$^*$ & 68.6 & 52.4 & 81.5 & 1284.9 & 57.6 & 49.1 & 26.3 & 58.4  \\
    CDPruner{\small\texttt{(NIPS25)}} & 73.6 & 57.0 & {57.9}$^*$ & 69.5 & 52.1 & 87.1 & 1353.0 & 59.1 & 49.6 & 27.8 & 60.4  \\
    \textbf{EADP}{\small\texttt{(Ours)}} & \textbf{74.0} & \textbf{58.2} & \textbf{59.3}$^*$ & \textbf{69.3} & \textbf{52.7} & \textbf{86.7} & \textbf{1347.0} & \textbf{59.4} & \textbf{50.1} & \textbf{30.6} & \textbf{60.9} \\
    \bottomrule
    \end{tabular}
  }

\end{table}

\noindent \textbf{Benchmarks.}
For LLaVA models (LLaVA-1.5 and LLaVA-NeXT), we evaluate on 10 image-based VQA benchmarks: VQAv2~\cite{VQAv2}, GQA~\cite{GQA}, VizWiz~\cite{gurari2018vizwizgrandchallengeanswering}, ScienceQA-IMG~\cite{SQA}, TextVQA~\cite{TextVQA}, POPE~\cite{li-etal-2023-evaluating}, MME~\cite{fu2025mmecomprehensiveevaluationbenchmark}, MMBench~\cite{liu2024mmbenchmultimodalmodelallaround}, MMBench-CN~\cite{liu2024mmbenchmultimodalmodelallaround}, and MM-Vet~\cite{yu2024mmvetevaluatinglargemultimodal}.
For Qwen-series advanced VLMs, we follow their standard evaluation protocols and report results on a diverse set of benchmarks. Specifically, Qwen2.5-VL is evaluated on 8 benchmarks: TextVQA, ChartQA~\cite{masry-etal-2022-chartqa}, AI2D~\cite{kembhavi2016diagramworthdozenimages}, OCRBench~\cite{Liu_2024OCRBench}, HallBench~\cite{Guan_2024_CVPR}, MME, MMBench, and MMBench-CN. 
For Qwen3-VL, we further include DocVQA~\cite{mathew2021docvqadatasetvqadocument} and InfoVQA~\cite{mathew2021infographicvqa}.
We also evaluate LLaVA-Video on 3 video benchmarks: LongVideoBench~\cite{LongVideoBench}, MVBench~\cite{MVBench}, and Video-MME~\cite{fu2025videommefirstevercomprehensiveevaluation}. All experiments use default settings and metrics; task details are provided in the supplementary material.

\noindent \textbf{Comparisons.}
We compare EADP with recent representative visual token pruning methods spanning different design philosophies, including FastV~\cite{chen2024fastv}, PyramidDrop~\cite{xing2025pyramiddropacceleratinglargevisionlanguage}, SparseVLM~\cite{zhang2024sparsevlm}, LLaVA-Prumerge~\cite{shang2025prumerge}, TRIM~\cite{song2025trim}, VisionZip~\cite{yang2025visionzip}, DART~\cite{wen2025dart}, DivPrune~\cite{alvar2025divprune}, HiPrune~\cite{liu2025hiprune} and CDPruner~\cite{zhang2025cdpruner}.

\noindent \textbf{Implementation Details.}
For image-based experiments on LLaVA, we build our implementation upon the official release. 
For video tasks, we evaluate using \texttt{lmms-eval} toolkit~\cite{zhang2024lmmsevalrealitycheckevaluation}. As for Qwen-series models, we rely on \texttt{VLMEvalKit}~\cite{duan2024vlmevalkit} to perform evaluation, following its official configuration to ensure fair comparison with prior work. All experiments are conducted on NVIDIA RTX 3090 GPUs.

\begin{table}[t]
  \centering
  \caption{\textbf{Performance comparison on LLaVA-NeXT-7B.} Avg. denotes the average performance across 9 benchmarks (excluding VizWiz). Boldface indicates the results of our method only.}

  \label{tab:llava-next}
  \resizebox{\linewidth}{!}{
    \begin{tabular}{l|cccccccccc|c}
    \toprule
    \textbf{Method} & \textbf{$\text{VQA}^\text{V2}$} & \textbf{GQA} & \textbf{VizWiz} & \textbf{$\text{SQA}^\text{IMG}$} & \textbf{$\text{VQA}^\text{Text}$} & \textbf{POPE} & \textbf{MME} & \textbf{$\text{MMB}^\text{EN}$} & \textbf{$\text{MMB}^\text{CN}$} & \textbf{MMVet} & \textbf{Avg.} \\
    \rowcolor{lightgray!75}\multicolumn{12}{c}{\textit{Upper Bound, All 2,880 Tokens} ($\mathbf{100\%}$)} \\
    \rowcolor{lightgray!25} LLaVA-NeXT-7B & 81.3 & 62.5 & 55.2/60.9* & 67.6 & 60.3 & 86.8 & 1510.9 & 65.8 & 57.3 & 40.0 & 66.3 \\
    \rowcolor{lightgray!75}\multicolumn{12}{c}{\textit{Retain 640 Tokens} \textcolor{Green}{($\downarrow\mathbf{77.8\%}$)}} \\
    FastV{\small\texttt{(ECCV24)}} & 76.6 & 58.4 & 54.9 & 67.2 & 57.5 & 80.0 & 1433.2 & 62.4 & 54.1 & 37.1 & 62.8 \\
    PDrop{\small\texttt{(CVPR25)}} & 78.7 & 59.5 & 53.8 & 66.9 & 57.1 & 83.2 & 1456.3 & 63.3 & 54.8 & 35.4 & 63.5 \\
    SparseVLM{\small\texttt{(ICML25)}} & 79.5 & 61.4 & 53.6 & 67.3 & 58.7 & 85.7 & 1464.8 & 64.8 & 57.1 & 35.7 & 64.8 \\
    PruMerge+{\small\texttt{(2024.05)}} & 78.0 & 61.1 & 57.9 & 67.4 & 55.2 & 85.5 & 1476.9 & 64.0 & 56.6 & 33.6 & 63.9 \\
    TRIM{\small\texttt{(COLING25)}} & 77.9 & 61.7 & 54.8 & 67.1 & 55.5 & 85.9 & 1473.5 & 66.0 & 55.9 & 36.9 & 64.5 \\
    VisionZip{\small\texttt{(CVPR25)}} & 79.2 & 61.4 & 57.1 & 67.5 & 58.5 & 86.2 & 1492.1 & 65.8 & 58.3 & 39.2 & 65.6 \\
    DART{\small\texttt{(EMNLP25)}} & 78.7 & 61.0 & 57.0 & 68.0 & 59.0 & 85.7 & 1444.9 & 65.1 & 57.5 & 37.3 & 64.9 \\
    DivPrune{\small\texttt{(CVPR25)}} & 79.3 & 61.9 & {60.6}$^*$ & 67.8 & 57.0 & 86.9 & 1469.7 & 65.8 & 57.3 & 38.0 & 65.3 \\
    CDPruner{\small\texttt{(NeurIPS25)}} & 79.9 & 62.3 & {60.8}$^*$ & 67.9 & 58.4 & 87.1 & 1474.5 & 66.2 & 57.5 & 40.7 & 66.0 \\
    HiPrune{\small\texttt{(AAAI26)}} & 78.8 & 60.7 & {61.2}$^*$ & 68.0 & 48.6 & 85.3 & 1475.3 & 67.0 & 58.8 & 38.4 & 64.4 \\
    \textbf{EADP}{\small\texttt{(Ours)}} & \textbf{80.0} & \textbf{62.7} & \textbf{60.6}$^*$ & \textbf{68.0} & \textbf{59.2} & \textbf{87.4} & \textbf{1494.7} & \textbf{66.5} & \textbf{57.9} & \textbf{40.1} & \textbf{66.3} \\
    \rowcolor{lightgray!75}\multicolumn{12}{c}{\textit{Retain 320 Tokens} \textcolor{Green}{($\downarrow\mathbf{88.9\%}$)}} \\
    FastV{\small\texttt{(ECCV24)}} & 63.4 & 52.1 & 51.3 & 66.3 & 52.9 & 59.9 & 1134.3 & 55.8 & 46.4 & 21.1 & 52.7 \\
    PDrop{\small\texttt{(CVPR25)}} & 67.7 & 53.3 & 49.7 & 66.5 & 50.7 & 63.5 & 1180.7 & 56.9 & 48.9 & 23.7 & 54.5 \\
    SparseVLM{\small\texttt{(ICML25)}} & 73.9 & 57.8 & 54.2 & 67.0 & 56.0 & 77.8 & 1383.5 & 63.3 & 53.4 & 31.9 & 61.1 \\
    PruMerge+{\small\texttt{(2024.05)}} & 74.8 & 58.4 & 57.7 & 67.5 & 54.2 & 79.1 & 1423.7 & 62.6 & 54.8 & 32.3 & 61.7 \\
    TRIM{\small\texttt{(COLING25)}} & 75.3 & 60.2 & 53.5 & 66.4 & 50.9 & 85.7 & 1445.8 & 62.1 & 52.2 & 33.5 & 62.1 \\
    VisionZip{\small\texttt{(CVPR25)}} & 76.3 & 59.3 & 56.2 & 67.3 & 56.4 & 82.9 & 1402.7 & 62.7 & 55.6 & 35.2 & 62.9 \\
    DART{\small\texttt{(EMNLP25)}} & 75.1 & 59.1 & 56.8 & 67.1 & 56.3 & 81.4 & 1433.5 & 63.8 & 55.4 & 36.3 & 62.9 \\
    DivPrune{\small\texttt{(CVPR25)}} & 77.2 & 61.1 & {60.1}$^*$ & 67.5 & 56.2 & 84.7 & 1423.3 & 63.9 & 55.7 & 34.8 & 63.6 \\
    CDPruner{\small\texttt{(NeurIPS25)}} & 77.9 & 61.6 & {59.9}$^*$ & 67.7 & 56.4 & 86.8 & 1453.0 & 64.1 & 55.7 & 37.9 & 64.5 \\
    HiPrune{\small\texttt{(AAAI26)}} & 74.4 & 57.6 & {61.3}$^*$ & 67.3 & 56.5 & 78.9 & 1406.8 & 64.2 & 57.0 & 34.7 & 62.3 \\
    \textbf{EADP}{\small\texttt{(Ours)}} & \textbf{78.6} & \textbf{62.2} & \textbf{60.4}$^*$ & \textbf{67.6} & \textbf{57.0} & \textbf{86.9} & \textbf{1491.3} & \textbf{64.6} & \textbf{55.9} & \textbf{39.4} & \textbf{65.2} \\
    \rowcolor{lightgray!75}\multicolumn{12}{c}{\textit{Retain 160 Tokens} \textcolor{Green}{($\downarrow\mathbf{94.4\%}$)}} \\
    PruMerge+{\small\texttt{(2024.05)}} & 69.1 & 55.6 & 57.2 & 66.7 & 50.9 & 73.7 & 1298.3 & 58.3 & 49.1 & 27.6 & 57.3 \\
    TRIM{\small\texttt{(COLING25)}} & 70.3 & 57.1 & 52.9 & 65.7 & 47.4 & 84.4 & 1281.6 & 61.2 & 45.4 & 30.1 & 58.4 \\
    VisionZip{\small\texttt{(CVPR25)}} & 70.8 & 55.8 & 55.5 & 67.7 & 53.4 & 76.0 & 1319.4 & 59.2 & 51.2 & 31.6 & 59.1 \\
    DART{\small\texttt{(EMNLP25)}} & 73.1 & 56.3 & 56.7 & 67.5 & 53.6 & 77.2 & 1332.8 & 61.5 & 53.3 & 32.0 & 60.1 \\
    DivPrune{\small\texttt{(CVPR25)}} & 75.0 & 60.2 & {60.7}$^*$ & 67.1 & 53.7 & 80.0 & 1376.3 & 62.3 & 53.4 & 32.4 & 61.4 \\
    CDPruner{\small\texttt{(NeurIPS25)}} & 76.4 & 60.6 & {59.7}$^*$ & 67.5 & 53.2 & 86.3 & 1402.3 & 63.6 & 53.8 & 35.0 & 62.9 \\
    HiPrune{\small\texttt{(AAAI26)}} & 67.3 & 53.7 & {59.5}$^*$ & 68.7 & 48.6 & 67.7 & 1200.7 & 59.8 & 50.7 & 29.2 & 56.2 \\
    \textbf{EADP}{\small\texttt{(Ours)}} & \textbf{77.0} & \textbf{61.6} & \textbf{60.2}$^*$ & \textbf{67.4} & \textbf{54.2} & \textbf{86.0} & \textbf{1419.5} & \textbf{63.4} & \textbf{54.6} & \textbf{35.6} & \textbf{63.4} \\
    \bottomrule
    \end{tabular}
  }
\end{table}

\subsection{Results on LLaVA series}

\noindent \textbf{Standard-resolution inputs.}
We evaluate EADP on LLaVA-1.5-7B and report results in \cref{tab:llava-1.5} under 3 common budgets of retained tokens. EADP consistently achieves the highest average accuracy across all pruning methods at each budget. With 77.8\% tokens pruned, it reaches 63.5 average, closest to the full-token upper bound, surpassing CDPruner and DivPrune. Under heavier pruning, it remains strong (62.2 at 64 tokens; 60.9 at 32 tokens), beating DivPrune by 2.5 points on average and showing robustness to extreme reduction. EADP also excels on GQA and MMVet, highlighting its fine-grained vision–instruction modeling.

\noindent \textbf{High-resolution inputs.}
While higher visual resolution improves fine-grained reasoning in VLMs~\cite{liu2023llava,bai2025qwen25vltechnicalreport,chen2024internvl2.5}, it also increases redundancy, making token pruning essential.
We evaluated EADP on LLaVA-NeXT-7B with $672 \times 672$ input resolution, producing 2880 visual tokens per image. As shown in \cref{tab:llava-next}, EADP is highly robust in this high-density regime: with 640 tokens, it matches the unpruned upper bound at 66.3 average. With 320/160 tokens, it achieves 65.2/63.4, respectively, outperforming strong recent baselines. These results confirm EADP’s ability to preserve fine-grained semantics under high-resolution inputs.

\subsection{Results on advanced VLMs}

\noindent \textbf{Qwen2.5-VL.}
We further evaluate EADP on the recent open-source VLM Qwen2.5-VL~\cite{bai2025qwen25vltechnicalreport} to assess generalization across heterogeneous architectures. Following CDPruner~\cite{zhang2025cdpruner}, we fix the input resolution to $1008 \times 1008$, yielding 1,296 visual tokens per image. As Qwen2.5-VL uses a different visual encoder and projection design, methods requiring a dedicated \texttt{[CLS]} token are inapplicable; thus we compare only with DivPrune~\cite{alvar2025divprune}, HiPrune~\cite{liu2025hiprune}, and CDPruner. Empirically, pruning causes larger degradation on Qwen2.5-VL than on LLaVA, suggesting its visual tokens are more tightly coupled with downstream reasoning. As shown in \cref{tab:qwen2.5-vl}, when retaining 256 tokens, EADP maintains 74.3 average accuracy, outperforming all baselines by a clear margin. Even under the highly compressed 128-token setting, EADP preserves 68.4 accuracy, demonstrating robustness under extreme token constraints. Specifically, EADP’s advantage is particularly evident on hallucination-sensitive and reasoning-intensive benchmarks such as HallBench.

\noindent \textbf{Qwen3-VL.}
We also evaluate EADP on Qwen3-VL-8B and additionally report results on DocVQA and InfoVQA, which require preserving fine-grained textual and layout information in document images. As shown in~\cref{tab:qwen3vl8b}, EADP consistently achieves best average performance under all token budgets. Specifically, when retaining 128 tokens, EADP obtains 62.7 average scores, outperforming the strongest baseline by 3.5. Gains are particularly evident on document-oriented benchmarks: at 256 tokens, EADP improves DocVQA by 7.0 and 9.8 over DivPrune and CDPruner, respectively, while also bringing improvements on InfoVQA. These results demonstrate that EADP can effectively preserve fine-grained textual evidence and holistic visual structures under strict token budgets.

\begin{table}[!t]
  \centering
  \caption{\textbf{Performance comparison on Qwen2.5-VL-7B.} Avg. denotes the average performance across 8 benchmarks. Boldface indicates the results of our method only.}
  \label{tab:qwen2.5-vl}
  \resizebox{\linewidth}{!}{
    \begin{tabular}{l|cccccccc|c}
    \toprule
    \textbf{Method} & \textbf{TextVQA} & \textbf{ChartQA} & \textbf{AI2D} & \textbf{OCRBench} & \textbf{HallBench} & \textbf{MME} & \textbf{MMB-EN} & \textbf{MMB-CN} & \textbf{Avg.} \\
    \rowcolor{lightgray!75}\multicolumn{10}{c}{\textit{Upper Bound, All 1296 Tokens} ($\mathbf{100\%}$)} \\
    \rowcolor{lightgray!25} Qwen2.5-VL-7B & 84.5 & 86.4 & 84.2 & 869 & 47.7 & 2303.4 & 83.9 & 82.8 & 84.0 \\
    \rowcolor{lightgray!75}\multicolumn{10}{c}{\textit{Retain 512 Tokens} \textcolor{Green}{($\downarrow\mathbf{60.5\%}$)}} \\
    DivPrune{\small\texttt{(CVPR25)}} & 78.8 & 77.1 & 79.6 & 760 & 42.2 & 2188.4 & 81.7 & 80.4 & 78.1 \\
    CDPruner{\small\texttt{(NeurIPS25)}} & 66.2 & 67.5 & 72.9 & 625 & 39.9 & 2122.4 & 79.6 & 78.3 & 71.6 \\
    HiPrune{\small\texttt{(AAAI26)}} & 75.8 & 74.2 & 79.3 & 679 & 40.5 & 2176.5 & 80.3 & 80.1 & 75.9 \\
    \textbf{EADP}{\small\texttt{(Ours)}} & \textbf{78.6} & \textbf{76.2} & \textbf{79.5} & \textbf{744} & \textbf{42.2} & \textbf{2213.3} & \textbf{81.6} & \textbf{80.8} & \textbf{78.0} \\
    \rowcolor{lightgray!75}\multicolumn{10}{c}{\textit{Retain 256 Tokens} \textcolor{Green}{($\downarrow\mathbf{80.2\%}$)}} \\
    DivPrune{\small\texttt{(CVPR25)}} & 74.0 & 67.0 & 77.4 & 665 & 36.4 & 2173.0 & 80.5 & 78.3 & 73.6 \\
    CDPruner{\small\texttt{(NeurIPS25)}} & 55.1 & 56.8 & 69.8 & 509 & 34.4 & 2017.6 & 77.6 & 74.7 & 65.0 \\
    HiPrune{\small\texttt{(AAAI26)}} & 64.2 & 56.7 & 74.1 & 579 & 37.3 & 2153.2 & 78.4 & 79.0 & 69.4 \\
    \textbf{EADP}{\small\texttt{(Ours)}} & \textbf{73.8} & \textbf{67.4} & \textbf{78.7} & \textbf{668} & \textbf{38.7} & \textbf{2202.0} & \textbf{80.2} & \textbf{78.5} & \textbf{74.3} \\
    \rowcolor{lightgray!75}\multicolumn{10}{c}{\textit{Retain 128 Tokens} \textcolor{Green}{($\downarrow\mathbf{90.1\%}$)}} \\
    DivPrune{\small\texttt{(CVPR25)}} & 66.1 & 51.1 & 73.7 & 516 & 31.2 & 2054.9 & 77.4 & 77.3 & 66.4 \\
    CDPruner{\small\texttt{(NeurIPS25)}} & 44.7 & 47.2 & 68.3 & 389 & 27.8 & 1876.7 & 73.3 & 71.7 & 58.2 \\
    HiPrune{\small\texttt{(AAAI26)}} & 51.1 & 41.3 & 72.3 & 497 & 31.2 & 2026.8 & 75.0 & 76.2 & 62.3 \\
    \textbf{EADP}{\small\texttt{(Ours)}} & \textbf{65.8} & \textbf{51.9} & \textbf{75.7} & \textbf{556} & \textbf{36.7} & \textbf{2137.9} & \textbf{78.4} & \textbf{76.4} & \textbf{68.4} \\
    \bottomrule
    \end{tabular}
  }
\end{table}

\begin{table}[h]
  \centering
  \caption{\textbf{Performance comparison on Qwen3-VL-8B.} Avg. denotes the average performance across 10 benchmarks. Boldface indicates the results of our method only.}
  \label{tab:qwen3vl8b}
  \resizebox{\textwidth}{!}{
    \begin{tabular}{l|cccccccccc|c}
    \toprule
    \textbf{Method} & \textbf{TextVQA} & \textbf{ChartQA} & \textbf{AI2D} & \textbf{OCRBench} & \textbf{HallBench} & \textbf{MME} & \textbf{MMB-EN} & \textbf{MMB-CN} & \textbf{DocVQA} & \textbf{InfoVQA} & \textbf{Avg.} \\
    \rowcolor{lightgray!75}\multicolumn{12}{c}{\textit{Upper Bound, All 1024 Tokens} ($\mathbf{100\%}$)} \\
    \rowcolor{lightgray!25} Qwen3-VL-8B & 83.8 & 82.3 & 86.3 & 855 & 51.2 & 2440.2 & 86.3 & 84.6 & 94.5 & 66.4 & 84.3 \\
    \rowcolor{lightgray!75}\multicolumn{12}{c}{\textit{Retain 512 Tokens} \textcolor{Green}{($\downarrow\mathbf{50.0\%}$)}} \\
    DivPrune{\small\texttt{(CVPR25)}} & 75.0 & 59.8 & 78.6 & 658 & 41.0 & 2217.5 & 82.1 & 79.4 & 74.6 & 46.0 & 71.3 \\
    CDPruner{\small\texttt{(NeurIPS25)}} & 74.5 & 62.0 & 76.9 & 660 & 41.6 & 2199.8 & 81.0 & 79.3 & 72.3 & 47.3 & 71.1 \\
    \textbf{EADP}{\small\texttt{(Ours)}} & \textbf{75.2} & \textbf{65.9} & \textbf{77.7} & \textbf{673} & \textbf{42.5} & \textbf{2261.1} & \textbf{81.6} & \textbf{79.7} & \textbf{77.7} & \textbf{49.9} & \textbf{73.1} \\
    \rowcolor{lightgray!75}\multicolumn{12}{c}{\textit{Retain 256 Tokens} \textcolor{Green}{($\downarrow\mathbf{75.0\%}$)}} \\
    DivPrune{\small\texttt{(CVPR25)}} & 69.5 & 47.5 & 75.2 & 595 & 38.7 & 2185.3 & 80.0 & 78.8 & 55.8 & 38.3 & 65.3 \\
    CDPruner{\small\texttt{(NeurIPS25)}} & 68.5 & 51.9 & 74.1 & 601 & 41.3 & 2200.6 & 79.8 & 80.0 & 53.0 & 37.6 & 65.6 \\
    \textbf{EADP}{\small\texttt{(Ours)}} & \textbf{71.4} & \textbf{55.5} & \textbf{75.7} & \textbf{625} & \textbf{42.8} & \textbf{2202.4} & \textbf{80.5} & \textbf{78.9} & \textbf{62.8} & \textbf{42.3} & \textbf{68.2} \\
    \rowcolor{lightgray!75}\multicolumn{12}{c}{\textit{Retain 128 Tokens} \textcolor{Green}{($\downarrow\mathbf{87.5\%}$)}} \\
    DivPrune{\small\texttt{(CVPR25)}} & 62.6 & 37.0 & 72.4 & 499 & 36.7 & 2136.3 & 77.0 & 76.6 & 39.5 & 33.8 & 59.2 \\
    CDPruner{\small\texttt{(NeurIPS25)}} & 60.6 & 40.4 & 71.7 & 515 & 34.7 & 2153.8 & 77.9 & 77.5 & 37.8 & 32.5 & 59.2 \\
    \textbf{EADP}{\small\texttt{(Ours)}} & \textbf{66.1} & \textbf{43.6} & \textbf{73.6} & \textbf{550} & \textbf{40.2} & \textbf{2200.2} & \textbf{79.9} & \textbf{78.4} & \textbf{45.3} & \textbf{35.3} & \textbf{62.7} \\
    \bottomrule
    \end{tabular}
  }
\end{table}

\subsection{Results on Video Understanding}

Video understanding poses a more challenging scenario for visual token efficiency, as redundancy naturally accumulates across both spatial and temporal dimensions~\cite{shao2025holitomholistictokenmerging}. To examine EADP in this high-redundancy setting, we conduct experiments on LLaVA-Video~\cite{zhang2025llavavideovideoinstructiontuning}, a VLM tailored for video reasoning. 
As shown in \cref{tab:llava-video}, EADP consistently achieves the best performance across all configuration. 
When the pruning ratio reaches 81.1\%, our method achieves 57.0 average accuracy, only 4.2 lower than original performance and outperforms all baselines. Importantly, the performance gap becomes more pronounced as the pruning ratio increases, suggesting that EADP better identifies temporally and semantically critical tokens, preventing severe information loss under high compression.
\emph{More detailed results are provided in the supplementary materials.}

\begin{table}[!t]
  \noindent
  \begin{minipage}[t]{0.516\textwidth}
    \centering
    \captionof{table}{Performance comparison on LLaVA-Video-7B with 64 frames per video.}

    \resizebox{\linewidth}{!}{%
    \begin{tabular}{l|ccc|c}
    \toprule
    \textbf{Method} & \textbf{MVBench} & \textbf{LongVideoBench} & \textbf{Video-MME} & \textbf{Avg.} \\
    \rowcolor{lightgray!75}\multicolumn{5}{c}{\textit{Upper Bound, All 64 $\times$ 169 Tokens} ($\mathbf{100\%}$)} \\
    \rowcolor{lightgray!25} LLaVA-Video-7B & 60.4 & 58.7 & 64.4 & 61.2 \\
    \rowcolor{lightgray!75}\multicolumn{5}{c}{\textit{Retain 64 $\times$ 64 Tokens} \textcolor{Green}{($\downarrow\mathbf{62.1\%}$)}} \\
    DivPrune{\small\texttt{(CVPR25)}} & 55.2 & 58.7 & 61.2 & 58.4 \\
    CDPruner{\small\texttt{(NeurIPS25)}} & 54.1 & 57.4 & 60.9 & 57.5 \\
    HiPrune{\small\texttt{(AAAI26)}} & 54.0 & 56.0 & 59.0 & 56.3 \\
    \textbf{EADP}{\small\texttt{(Ours)}} & \textbf{55.7} & \textbf{57.9} & \textbf{62.0} & \textbf{58.5} \\
    \rowcolor{lightgray!75}\multicolumn{5}{c}{\textit{Retain 64 $\times$ 32 Tokens} \textcolor{Green}{($\downarrow\mathbf{81.1\%}$)}} \\
    DivPrune{\small\texttt{(CVPR25)}} & 53.7 & 55.7 & 59.2 & 56.2 \\
    CDPruner{\small\texttt{(NeurIPS25)}} & 53.2 & 55.4 & 58.3 & 55.6 \\
    HiPrune{\small\texttt{(AAAI26)}} & 51.1 & 54.9 & 56.0 & 54.0 \\
    \textbf{EADP}{\small\texttt{(Ours)}} & \textbf{54.5} & \textbf{56.6} & \textbf{60.4} & \textbf{57.2} \\
    \rowcolor{lightgray!75}\multicolumn{5}{c}{\textit{Retain 64 $\times$ 16 Tokens} \textcolor{Green}{($\downarrow\mathbf{90.5\%}$)}} \\
    DivPrune{\small\texttt{(CVPR25)}} & 52.1 & 52.7 & 57.2 & 54.0 \\
    CDPruner{\small\texttt{(NeurIPS25)}} & 50.2 & 53.8 & 56.3 & 53.4 \\
    HiPrune{\small\texttt{(AAAI26)}} & 48.8 & 52.0 & 53.8 & 51.5 \\
    \textbf{EADP}{\small\texttt{(Ours)}} & \textbf{52.6} & \textbf{54.5} & \textbf{57.3} & \textbf{54.8} \\
    \bottomrule
    \end{tabular}
    }
    \label{tab:llava-video}
  \end{minipage}%
  \hfill
  \begin{minipage}[t]{0.47\textwidth}
    \centering
    \captionof{table}{Efficiency analysis on LLaVA-1.5-7B.}

    \resizebox{\linewidth}{!}{
    \begin{tabular}{l|ccc}
    \toprule
    \textbf{Method} & \textbf{Prefill(ms)} & \textbf{Latency(ms)} & \textbf{FLOPs(G)}\\
    \rowcolor{lightgray!75}\multicolumn{4}{c}{\textit{Upper Bound, All 576 Tokens} ($\mathbf{100\%}$)} \\
    \rowcolor{lightgray!25} LLaVA-1.5-7B & 207.4 & 256.8 & 4489.8 \\
    \rowcolor{lightgray!75}\multicolumn{4}{c}{\textit{Retain 128 Tokens} \textcolor{Green}{($\downarrow\mathbf{77.8\%}$)}} \\
    DivPrune{\small\texttt{(CVPR25)}} & 109.6 ($\times$1.9) & 158.0 ($\times$1.6) & 1529.9 ($\times$2.9) \\
    CDPruner{\small\texttt{(NeurIPS25)}} & 135.9 ($\times$1.5) & 183.3 ($\times$1.4) & 1538.3 ($\times$2.9) \\
    HiPrune{\small\texttt{(AAAI26)}} & 99.4 ($\times$2.0) & 150.1 ($\times$1.7) & 1529.9 ($\times$2.9) \\
    \textbf{EADP}{\small\texttt{(Ours)}} & \textbf{118.8 ($\times$1.7)} & \textbf{169.3 ($\times$1.5)} & \textbf{1538.4 ($\times$2.9)} \\
    \rowcolor{lightgray!75}\multicolumn{4}{c}{\textit{Retain 64 Tokens} \textcolor{Green}{($\downarrow\mathbf{88.9\%}$)}} \\
    DivPrune{\small\texttt{(CVPR25)}} & 92.7 ($\times$2.2) & 140.4 ($\times$1.8) & 1107.0 ($\times$4.1) \\
    CDPruner{\small\texttt{(NeurIPS25)}} & 105.9 ($\times$2.0) & 157.3 ($\times$1.6) & 1115.5 ($\times$4.0) \\
    HiPrune{\small\texttt{(AAAI26)}} & 85.7 ($\times$2.4) & 137.6 ($\times$1.9) & 1107.0 ($\times$4.1) \\
    \textbf{EADP}{\small\texttt{(Ours)}} & \textbf{97.0 ($\times$2.1)} & \textbf{147.7 ($\times$1.7)} & \textbf{1115.6 ($\times$4.0)} \\
    \rowcolor{lightgray!75}\multicolumn{4}{c}{\textit{Retain 32 Tokens} \textcolor{Green}{($\downarrow\mathbf{94.4\%}$)}} \\
    DivPrune{\small\texttt{(CVPR25)}} & 79.8 ($\times$2.6) & 128.3 ($\times$2.0) & 895.6 ($\times$5.0) \\
    CDPruner{\small\texttt{(NeurIPS25)}} & 85.8 ($\times$2.4) & 137.46 ($\times$1.9) & 904.1 ($\times$5.0) \\
    HiPrune{\small\texttt{(AAAI26)}} & 74.8 ($\times$2.8) & 126.54 ($\times$2.0) & 895.6 ($\times$5.0) \\
    \textbf{EADP}{\small\texttt{(Ours)}} & \textbf{81.9 ($\times$2.5)} & \textbf{129.63 ($\times$2.0)} & \textbf{904.1 ($\times$5.0)} \\
    \bottomrule
    \end{tabular}
    }
    
    \label{tab: efficiency}
  \end{minipage}

\end{table}

\subsection{Efficiency Analysis}

We evaluate the computational efficiency of EADP in terms of prefill time, end-to-end inference latency and FLOPs. All metrics are averaged across 2,000 instances equally sampled from VizWiz, TextVQA, POPE, and MME. 
\cref{tab: efficiency} presents our main results on LLaVA-1.5-7B under three visual token budgets. \emph{Additional results on different architectures are detailed in the supplementary material.}
Compared to the unpruned LLaVA-1.5-7B, EADP consistently delivers significant acceleration. For instance, retaining 128 tokens reduces FLOPs by nearly 66\% while achieving a $1.75 \times$ speedup in prefill time and a $1.50 \times$ speedup in end-to-end latency.
When compared to existing approaches such as DivPrune and HiPrune, EADP exhibits a marginal computational overhead which originates from the dense scoring. However, considering the superior performance, this minimal overhead represents a highly worthwhile trade-off. \emph{We further report a wall-clock time analysis for the different components of our method in supplementary material.}

\vspace{-1mm}
\subsection{Ablation Studies}
\begin{table*}[!t]
\centering
\caption{\textbf{Ablation studies on core design choices.} We evaluate the impact of various hyperparameters and algorithmic variants across five benchmarks. All experiments are conducted using LLaVA-1.5-7B with a fixed budget of 128 visual tokens.}
\vspace{-1mm}
\label{tab:ablation-128}

\begin{minipage}[t]{0.49\textwidth}

    \centering
    \resizebox{\linewidth}{!}{
    \begin{tabular}{lcccccc}
    \toprule
     & \textbf{VizWiz} & \textbf{SQA} & \textbf{TextVQA} & \textbf{POPE} & \textbf{MME} & \textbf{Avg.} \\
    \rowcolor{lightgray!75}\multicolumn{7}{c}{global-dense balancing coefficient} \\
    $\mu=$0.0 & 57.8 & 68.8 & 56.2 & 84.6 & 1421.9 & 67.7 \\
    $\mu=$0.2 & 58.2 & 68.9 & 56.3 & 85.9 & 1422.0 & 68.1 \\
    $\mu=$0.5 & 57.9 & 69.0 & 56.5 & 87.2 & 1439.0 & 68.5 \\
    $\mu=$0.8 & 58.0 & 68.9 & 56.5 & 86.4 & 1440.6 & 68.3 \\
    $\mu=$1.0 & 57.5 & 68.7 & 56.1 & 87.3 & 1422.2 & 68.1 \\
    \rowcolor{lightgray!75}\multicolumn{7}{c}{keep proportion} \\
    $q=$0.3 & 58.2 & 69.0 & 56.6 & 87.2 & 1437.2 & 68.6 \\
    $q=$0.5 & 58.0 & 69.0 & 56.5 & 87.2 & 1439.0 & 68.5 \\
    $q=$0.7 & 57.4 & 68.3 & 56.0 & 86.7 & 1420.1 & 67.9 \\
    $q=$0.9 & 57.1 & 68.2 & 55.8 & 86.4 & 1411.0 & 67.6 \\
    \rowcolor{lightgray!75}\multicolumn{7}{c}{aggregation method} \\
    \textbf{(1)} & 58.0 & 69.0 & 56.5 & 87.2 & 1439.0 & 68.5 \\
    \textbf{(2)} & 57.1 & 68.3 & 55.9 & 87.0 & 1430.2 & 68.0 \\
    \textbf{(3)} & 58.0 & 69.2 & 56.5 & 87.2 & 1434.3 & 68.5 \\
    \textbf{(4)} & 58.3 & 68.9 & 56.3 & 87.4 & 1441.9 & 68.6 \\
    
    \bottomrule
    \end{tabular}
    }
\end{minipage}
\hfill
\begin{minipage}[t]{0.49\textwidth}

    \centering
    \resizebox{1.08\linewidth}{!}{
    \begin{tabular}{lcccccc}
    \toprule
     & \textbf{VizWiz} & \textbf{SQA} & \textbf{TextVQA} & \textbf{POPE} & \textbf{MME} & \textbf{Avg.} \\
    \rowcolor{lightgray!75}\multicolumn{7}{c}{polarization value} \\
    $\beta=$0.0 & 58.0 & 68.8 & 55.9 & 86.7 & 1395.4 & 67.8 \\
    $\beta=$0.5 & 57.9 & 69.1 & 56.0 & 87.0 & 1394.9 & 67.9 \\
    $\beta=$2.0 & 57.8 & 69.1 & 56.5 & 87.2 & 1412.4 & 68.2 \\
    $\beta=$5.0 & 58.0 & 69.0 & 56.6 & 87.3 & 1414.1 & 68.3 \\
    $\beta=$10.0 & 56.8 & 68.8 & 56.4 & 87.2 & 1439.0 & 68.2 \\
    \rowcolor{lightgray!75}\multicolumn{7}{c}{spatial smoothing size} \\
    \textbf{w.o.}& 57.9 & 68.6 & 55.7 & 86.9 & 1413.4 & 67.9 \\
    size=3 & 58.0 & 69.0 & 56.5 & 87.2 & 1439.0 & 68.5 \\
    size=5 & 57.9 & 69.2 & 56.5 & 87.2 & 1435.0 & 68.5 \\
    size=7 & 57.9 & 69.1 & 56.3 & 87.2 & 1435.3 & 68.4 \\
    \rowcolor{lightgray!75}\multicolumn{7}{c}{how to measure dispersion degree} \\
    \textbf{(A)} & 58.0 & 69.0 & 56.5 & 87.2 & 1439.0 & 68.5 \\
    \textbf{(B)} & 57.9 & 69.3 & 56.4 & 87.2 & 1429.8 & 68.4 \\
    \textbf{(C)} & 57.9 & 69.2 & 56.4 & 87.3 & 1435.1 & 68.5 \\
    \bottomrule
    \end{tabular}
    }
\end{minipage}
\vspace{-2mm}
\end{table*}

\vspace{-1mm}

\noindent \textbf{Ablation on the global-dense balancing coefficient.}
The coefficient $\mu$ balances the global and dense guidance score. Relying purely on dense guidance score ($\mu = 0.0$) yields the lowest performance. As we increase $\mu$, the performance steadily improves and peak at $\mu = 0.5$. 
This trend demonstrates that global scene representation provides an indispensable foundation for holistic understanding, while fine-grained dense alignment serves as a crucial supplement for details.

\noindent \textbf{Ablation on the polarization value.}
The hyperparameter $\beta$ is designed to sharpen the score map, distinctively prioritizing salient region.
Disabling polarization ($\beta = 0.0$) significantly degrades performance. Interestingly, the performance reaches a plateau when $\beta \geq 1.0$, indicating that the submodular maximization is highly robust.

\noindent \textbf{Ablation on the keep proportion.}
We analyze the impact of preserving only a proportion $q$ of text tokens with lowest entropy while calculating the dense guidance score. Notably, aggressively discarding 70\% of text tokens ($q = 0.3$) achieves the highest score, which strongly corroborates our core motivation: high-entropy text tokens correspond to dispersed, uninformative visual attention. 

\noindent \textbf{Ablation on spatial smoothing size.}
Spatial smoothing is crucial for obtaining contiguous scores maps. Compared to the case applying a $3 \times 3$ or $5 \times 5$ kernel, removing the smoothing ("w.o.") leads to a distinct performance degradation. 
This confirms that enforcing spatial coherence is essential, which could mitigate the impact of localized noise and guide the submodular optimizer to select structurally complete semantic regions.

\noindent \textbf{Ablation on aggregation method.} For weighting the semantic relevance scores of the filtered texts (see~\cref{eq:aggregation_method}), we consider four styles: (1) using the weighted inverse entropy, (2) averaging over the top-k scores, (3) using the gated L1 norm, and (4) our proposed method. As shown in~\cref{tab:ablation-128}, after removing noisy texts, effectively weighting each visual score computed from the remaining text tokens is beneficial, whereas the simple mean operation (Style 2) yields smaller gains than the other three. For simplicity and robustness, we ultimately adopt style 4.

\noindent \textbf{Ablation on dispersion measurement.} We explored three statistics to quantify the degree of dispersion: (A) entropy, (B) central mass ratio, and (C) variance. As shown in~\cref{tab:ablation-128}, we found that these schemes achieve comparable performance, demonstrating that the noise-filtering process is robust to the choice of statistic. \emph{More details of the ablation studies can be referred to the supplementary material.}

\section{Conclusion}
\label{sec:conclusion}
In this work, we investigate the failure modes of visual token pruning in VLMs and identify textual noise dispersion and feature fragmentation as critical bottlenecks. To address these issues, we introduce EADP, which reformulates visual pruning as a structured compression task. EADP elegantly quantifies and filters textual noise via statistical entropy to yield robust instruction relevance scores. Furthermore, by casting token selection as a facility location submodular maximization problem with spatial priors, EADP explicitly guarantees holistic and non-redundant visual coverage. Extensive experiments across diverse VLM architectures demonstrate that EADP achieves state-of-the-art performance, robustly preserving fine-grained visual cues under strict token budgets and significantly advancing the accuracy-efficiency trade-off for practical VLM deployment.
\vspace{-1mm}

\section*{Acknowledgements}

This work was supported by the NSFC under Grant 62322604 and 62576207.


\title{Combating Textual Noise and Redundancy: Entropy-Aware Dense Visual Token Pruning - Supplementary Materials}
\titlerunning{Supplementary Materials}

\author{Xuehui Wang\inst{}$^{\dagger}$ \orcidlink{0000-0002-6333-7773} \and
Xuankun Yang\inst{}$^{\dagger}$ \orcidlink{0009-0001-0763-5776} \and
Wei Shen\inst{}$^\text{\faEnvelope}$ \orcidlink{0000-0002-1235-598X}}

\def\customsymbol#1{
    \ifcase\number\value{#1}
        \or*
        \or\faEnvelope
    \else\@ctrerr
    \fi
}
\authorrunning{X.~Wang et al.}
\institute{Shanghai Jiao Tong University, Shanghai, China \\
\email{\{wangxuehui,kk-dao,wei.shen\}@sjtu.edu.cn}\\
Codes: \url{https://github.com/SJTU-DeepVisionLab/EADP}
}

\maketitle
\section{Overview}
\label{sec:supp_overview}

In the supplementary material, we primarily provide additional experiments and analyzes, organized as follows:
\begin{itemize}
    \item In~\cref{sec:supp_proof}, we first provide detailed derivations to justify the formulation and efficiency of the submodular optimization problem
    \item In~\cref{sec:supp_implementation}, we present additional implementation details of the overall method. 
    \item In~\cref{sec:supp_experiment}, we describe the benchmarks used in our experiments and clarify the model architectures and variants considered. 
    \item In~\cref{sec:supp_more}, we analyze performance on more models and benchmarks to further validate the generalization of our approach. 
    \item In~\cref{sec:supp_more_ablation}, we include more ablations to document the hyperparameter selection process, along with detailed descriptions for our choice of aggregation method in our main paper.
\end{itemize}

\section{Theoretical Analysis of the Greedy Strategy}
\label{sec:supp_proof}

In this section, we provide a theoretical analysis of our method. We formulate the token selection process as a Facility Location Problem and prove that our objective function is both monotone and submodular. Consequently, the greedy selection strategy employed in our approach guarantees a $(1 - 1/e)$-approximation to the optimal solution.

\subsection{Problem Formulation}
Let $\mathcal{V} = \{v_1, v_2, \dots, v_N\} \in \mathbb{R}^{N \times d_v}$ denote the set of all visual tokens. Our goal is to select a subset $\mathcal{Y} \subseteq \mathcal{V}$ with a cardinality constraint $|\mathcal{Y}| \leq K \ll N$ (where $K$ is the target number of preserved tokens) that maximizes the representation of the original semantic information. The optimization objective is defined as:

\begin{equation}
    \label{eq:objective}
    \arg\max_{\mathcal{Y} \subseteq \mathcal{V}, |\mathcal{Y}| \leq K} \sum_{j=1}^{N} \hat{s}^I_j \cdot \max_{v_i\in\mathcal{Y}} \text{Sim}(v_i, v_j)
\end{equation}

where $\hat{s}^I_j$ represents the instruction relevance score of token $v_j$, and $\text{Sim}(v_i, v_j)$ denotes the cosine similarity between the visual features of token $v_i$ and $v_j$. This formulation aligns with the classic Metric Facility Location problem, where the selected tokens in $\mathcal{Y}$ serve as facilities to cover the clients in $\mathcal{V}$.

\subsection{Properties of the Objective Function}
To establish the approximation guarantee of the greedy algorithm, we analyze two key properties of the objective function $F(S) = \sum_{j=1}^{N} \hat{s}^I_j \cdot \max_{v_i\in S} \text{Sim}(v_i, v_j)$: \textit{monotonicity} and \textit{submodularity}.

\subsubsection{Monotonicity}
A set function $F: 2^\mathcal{V} \rightarrow \mathbb{R}$ is monotone if for every $A \subseteq B \subseteq \mathcal{V}$, $F(A) \leq F(B)$.

\textbf{Proof.} Consider the term for a single token $v_j$: $g_j(S) = \max_{v_i \in S} \text{Sim}(v_i, v_j)$.
For any sets $A \subseteq B$, the set of similarity values $\{\text{Sim}(v_i, v_j) \mid v_i \in A\}$ is a subset of $\{\text{Sim}(v_i, v_j) \mid v_i \in B\}$. Since the maximum operator is non-decreasing with respect to set inclusion, we have:
\begin{equation}
    \max_{v_i \in A} \text{Sim}(v_i, v_j) \leq \max_{v_i \in B} \text{Sim}(v_i, v_j) \implies g_j(A) \leq g_j(B)
\end{equation}
Instruction relevance score $S^I$ is applied to \texttt{min-max} normalization, and the subsequent smoothing and polarization operations retain the non-negativity property. 
Thus, score $\hat{s}^I_j$ is non-negative, the linear combination $F(S) = \sum_{j=1}^N \hat{s}^I_j \cdot g_j(S)$ preserves the inequality. Then we have $F(A) \leq F(B)$, and $F$ is monotone. \qed

\subsubsection{Submodularity}
A set function $F$ is submodular if it satisfies the property of diminishing returns. Formally, for every $A \subseteq B \subseteq \mathcal{V}$ and an element $u \in \mathcal{V} \setminus B$, the inequality holds:
\begin{equation}
    F(A \cup \{u\}) - F(A) \geq F(B \cup \{u\}) - F(B)
\end{equation}

\textbf{Proof.} Let us denote the marginal gain of adding token $u$ to a set $S$ for a specific target $v_j$ as $\Delta_j(u | S) = g_j(S \cup \{u\}) - g_j(S)$.
The term $g_j(S) = \max_{v_i \in S} \text{Sim}(v_i, v_j)$ represents the current best coverage for token $v_j$ by the set $S$. Let $c_j(S) = \max_{v_i \in S} \text{Sim}(v_i, v_j)$.
The gain can be rewritten as:
\begin{equation}
    \Delta_j(u | S) = \max(\text{Sim}(u, v_j), c_j(S)) - c_j(S) = \max(0, \text{Sim}(u, v_j) - c_j(S))
\end{equation}
Since $A \subseteq B$, we know that $c_j(A) \leq c_j(B)$ by monotonicity. A larger current coverage value implies a smaller (or equal) potential for improvement. Specifically:
\begin{equation}
    \text{Sim}(u, v_j) - c_j(A) \geq \text{Sim}(u, v_j) - c_j(B)
\end{equation}
Applying the $\max(0, \cdot)$ function, which is non-decreasing, we obtain:
\begin{equation}
    \max(0, \text{Sim}(u, v_j) - c_j(A)) \geq \max(0, \text{Sim}(u, v_j) - c_j(B))
\end{equation}
Therefore, $\Delta_j(u | A) \geq \Delta_j(u | B)$.
Since the total objective $F(S)$ is a non-negative linear combination of $g_j(S)$, the inequality sums over all $j$:
\begin{equation}
    F(A \cup \{u\}) - F(A) = \sum_{j=1}^N \hat{s}^I_j \cdot \Delta_j(u | A) \geq \sum_{j=1}^N \hat{s}^I_j \cdot \Delta_j(u | B) = F(B \cup \{u\}) - F(B)
\end{equation}
Thus, $F(S)$ is submodular. \qed

\subsection{Approximation Guarantee}
The problem of maximizing a monotone submodular function under a cardinality constraint is known to be NP-hard. However, the seminal work by Nemhauser et al. \cite{Nemhauser1978AnAO} proves that a greedy algorithm, which iteratively selects the element $u$ that maximizes the marginal gain $\Delta F(u | S)$, provides a tight approximation bound.

Specifically, let $\mathcal{Y}_{greedy}$ be the set selected by the greedy strategy and $\mathcal{Y}_{opt}$ be the optimal set. The following inequality holds:
\begin{equation}
    F(\mathcal{Y}_{greedy}) \geq \left(1 - \frac{1}{e}\right) F(\mathcal{Y}_{opt}) \approx 0.632 \cdot F(\mathcal{Y}_{opt})
\end{equation}

In our implementation, the calculation of the gain for each step:
\begin{equation}
    \text{Gain}(u) = \sum_{j = 1}^N \hat{s}^I_j \times (\max(\text{Sim}(u, v_j), \text{Curr}(v_j)) - \text{Curr}(v_j))
\end{equation}
corresponds exactly to the marginal gain maximization required by the theorem. Therefore, our pruning method is theoretically guaranteed to produce a near-optimal subset of tokens.

\section{Implementation Details}
\label{sec:supp_implementation}

In this section, we provide the specific implementation details in EADP. Algorithm \ref{alg:pruning} summarizes the overall streamlined pipeline.

\noindent\textbf{Cross-Modal Similarity Computation.}
It is worth noting that our calculation of the cross-modal similarity between text and visual tokens deviates from the conventional dot-product approach. Instead of utilizing the standard cosine similarity, we compute the \textbf{negative} cosine similarity. Although this inversion appears counter-intuitive at first glance, it fundamentally addresses a well-documented representational flaw in standard vision-language models. As revealed by ECLIP~\cite{li2022exploringvisualinterpretabilitycontrastive}, models pre-trained via global contrastive learning (e.g., CLIP~\cite{radford2021clip}) suffer from a ``semantic shift'' caused by their global pooling mechanisms. Consequently, their raw feature maps systematically and erroneously assign higher similarity scores to irrelevant background regions rather than the semantically meaningful foregrounds. By negating the cosine similarity, we effectively invert this biased attention distribution, enabling our method to accurately localize the critical foreground tokens. The necessity and empirical effectiveness of this crucial adjustment have also been corroborated by recent visual token prune approaches, such as CDPruner~\cite{zhang2025cdpruner} and TRIM~\cite{song2025trim}.

\begin{algorithm}[!t]
\caption{\textbf{Implementation of EADP.} \textit{In LLaVA-NEXT, $B = 5$, represents 5 patches; otherwise, $B = 1$ always holds.}}
\label{alg:pruning}
\KwIn{Visual tokens $\mathcal{V} \in \mathbb{R}^{B \times N \times d_v}$, Text embeddings $\mathcal{T} \in \mathbb{R}^{B \times (M + 1) \times d}$, Total Token Budget $K \in \mathbb{R}^B$}
\KwOut{Boolean token keep mask $\mathcal{M} \in \mathbb{R}^{B \times N}$}

\BlankLine
$S_{vv} \in \mathbb{R}^{B \times N \times N} \leftarrow \mathrm{ComputeVisualSimilarity}(\mathcal{V})$\;
$S^I \in \mathbb{R}^{B \times N} \leftarrow \mathrm{ComputeInstructionRelevanceScore}(\mathcal{V}, \mathcal{T})$\;
$S^I \leftarrow \frac{S^I - \min(S^I)}{\max(S^I) - \min(S^I) + \epsilon}$\;
$S^I_{\text{smooth}} \in \mathbb{R}^{B \times N} \leftarrow \mathrm{GaussianConv2D}(S^I, \text{kernel\_size}=3, \sigma=1.0)$\;
$\hat{S}^I \leftarrow \left(S^I_{\text{smooth}}\right)^\beta$\;

$Q \in \mathbb{R}^B = [Q_1, Q_2, \dots, Q_B] = \mathrm{DynamicQuotaAllocation}(S^I)$\;

Initialize selected token mask $\mathcal{M} \leftarrow \textbf{0} \in \mathbb{R}^{B \times N}$\;
Initialize max coverage $C \leftarrow \textbf{0} \in \mathbb{R}^{B \times N}$\;
$T_{max} \in \mathbb{R} \leftarrow \max(Q)$

\For{$t = 1, 2, \dots, T_{max}$}
{
    \tcp{Compute marginal gain for all candidate tokens per batch}
    $G_{b, u} \leftarrow \sum_{j=1}^{N} \hat{S}^I_{b, j} \cdot \max(0, S_{vv}[b, u, j] - C_{b, j}) \quad \forall b \in \{1..B\}, \forall u \in \{1..N\}$\;
    $G_{b, u} \leftarrow -\infty \quad \forall (b, u) \text{ where } \mathcal{M}_{b, u} = 1$\; 
    
    \tcp{Greedily select the token with the maximum gain per batch}
    $u^*_b \leftarrow \arg\max_{u} G_{b, u} \quad \forall b \in \{1..B\}$\;
    
    \tcp{Update maximum coverage state and boolean mask}
    $C_{b, j} \leftarrow \max(C_{b, j}, S_{vv}[b, u^*_b, j]) \quad \forall b \in \{1..B\}, \forall j \in \{1..N\}$\;
    $\mathcal{M}_{b, u^*_b} \leftarrow 1 \quad \forall b \text{ where } t \leq Q_b$\; 
}
\Return $\mathcal{M}$\;

\end{algorithm}

\noindent\textbf{Visual Similarity Rebound.}
The pairwise visual similarity is computed using cosine similarity between visual tokens, which naturally spans the range $[-1, 1]$. Since our submodular maximization objective requires non-negative marginal gains to ensure monotonic coverage expansion, we linearly project the cosine similarity to a strictly non-negative range $[0, 1]$. Specifically, the rebounded similarity matrix is computed as
\begin{equation}
    \text{Sim}(v_i, v_j) \leftarrow 0.5 \times (\text{Sim}(v_i, v_j) + 1).
\end{equation}
This stable range prevents negative interference during the greedy facility location process.

\noindent\textbf{Min-Max Normalization.}
Before applying spatial operations, the fused instruction relevance score $S^I$ are \texttt{min-max} normalized to a standard $[0, 1]$ scale. This guarantees that the importance distribution is strictly bounded and scale-invariant across different patches, ensuring stable subsequent power transformations.

\noindent\textbf{Spatial Smoothing.}
We reshape the instruction relevance score $S^I$ back to its correspongding 2D spatial map $S_{\text{2D}}^I \in \mathbb{R}^{H \times W}$ and apply a 2D Gaussian convolution with a kernel size of $3 \times 3$ and a standard deviation $\sigma = 1.0$. \texttt{Reflection padding} is utilized to mitigate boundary artifacts.

\noindent\textbf{Dynamic Quota Allocation.}
High-resolution MLLMs like LLaVA-NeXT employ an ``AnyRes'' strategy, which typically splits a high-resolution image into 5 independent patches: one downsampled global patch (denoted as $p=0$) and four unrolled local patches ($p \in \{1, 2, 3, 4\}$) corresponding to the top-left, top-right, bottom-left, and bottom-right regions. Since semantic information is unevenly distributed across the image, uniformly pruning these patches is sub-optimal. 

\noindent Given a total token budget $K$ for a single image, and letting $\hat{S}^I_{p, i}$ denote the instruction relevance score of the $i$-th token in patch $p$, we introduce two dynamic token allocation strategies to determine the exact quota $Q_p$ for each patch:

\begin{itemize}
    \item \textbf{Importance-Based Allocation:} This strategy calculates the cumulative instruction relevance score of each patch independently and allocates the budget proportionally. The weight $w_p$ for patch $p$ is defined as the sum of its token scores:
    \begin{equation}
        w_p = \sum_{i=1}^{N_p} \hat{S}^I_{p, i}, \quad \forall p \in \{0, 1, 2, 3, 4\}
    \end{equation}
    where $N_p$ is the total number of visual tokens in a single patch. The quota $Q_p$ is then dynamically assigned as:
    \begin{equation}
        Q_p = \max\left(1, \left\lfloor \frac{w_p}{\sum_{k=0}^{4} w_k} \cdot K \right\rfloor \right)
    \end{equation}
    ensuring that every patch retains at least 1 token to avoid structural corruption.
    
    \item \textbf{Global-Guided Allocation:} Alternatively, we leverage the global patch ($p=0$) as a macro-level semantic guide. The global patch sequence is reshaped into a 2D spatial grid $\hat{S}^{(0)}_{\text{2D}} \in \mathbb{R}^{H \times W}$ (where $H \times W = N_p$). We conceptually divide $\hat{S}^{(0)}_{\text{2D}}$ into $2 \times 2$ quadrants $\mathcal{A}_p$, explicitly corresponding to the spatial layout of the four local patches $p \in \{1, 2, 3, 4\}$. 
    
    To maintain a holistic view, we reserve a fixed $20\%$ of the budget for the global patch. The remaining $80\%$ is distributed among the local patches proportional to the relevance scores accumulated within their corresponding quadrants in the global guide. The guiding weight $w_p$ for local patch $p$ is computed as:
    \begin{equation}
        w_p = \sum_{(x,y) \in \mathcal{A}_p} \hat{S}^{(0)}_{\text{2D}, x,y}, \quad \forall p \in \{1, 2, 3, 4\}
    \end{equation}
    The allocated quotas are then formulated as:
    \begin{equation}
        Q_0 = \lfloor 0.2 K \rfloor, \quad 
        Q_p = \max\left(1, \left\lfloor \frac{w_p}{\sum_{k=1}^{4} w_k} \cdot (0.8 K) \right\rfloor \right) \quad \forall p \in \{1, 2, 3, 4\}
    \end{equation} 
\end{itemize}

\textbf{In all experiments involving LLaVA-NEXT in this paper, we use the first approach by default.}

\section{Details of Experimental Setup}
\label{sec:supp_experiment}

\subsection{Benchmarks}
In this section, we provide detailed descriptions of the datasets used in our experiments. We categorize these datasets into five groups: General Visual Question Answering, Text-oriented and Scientific Understanding, Comprehensive Multimodal Benchmarks, Hallucination Evaluation, and Video Understanding.

\subsubsection{General Visual Question Answering}
\begin{itemize}
    \item \textbf{VQAv2}~\cite{VQAv2}: VQAv2 is a widely used benchmark for Visual Question Answering. It is designed to counter the strong language priors found in its predecessor (VQA v1) by balancing the dataset such that for every question, there are complementary images that result in different answers. This forces models to rely on visual content rather than language statistics to answer correctly.
    
    \item \textbf{GQA}~\cite{GQA}: The GQA dataset focuses on visual reasoning and compositional question answering. It leverages scene graphs to generate questions that require multi-step inference, diverse reasoning skills (e.g., spatial, logical, relational), and a deep understanding of the visual scene structure.
    
    \item \textbf{VizWiz}~\cite{gurari2018vizwizgrandchallengeanswering}: VizWiz is a real-world VQA dataset originating from blind and visually impaired users. The images are often of lower quality (blur, poor lighting), and the questions reflect practical, daily needs. This dataset challenges models to handle imperfect visual data and recognize when a question cannot be answered based on the image content. \textbf{In our experiments, since the official challenge had expired, we used the validation split for evaluation.}
\end{itemize}

\subsubsection{Text-oriented and Scientific Understanding}
\begin{itemize}
    \item \textbf{TextVQA}~\cite{TextVQA}: TextVQA requires models to read and reason about text present within images to answer questions. It evaluates the Optical Character Recognition (OCR) capabilities of Multimodal Large Language Models (MLLMs) in natural scenes.
    
    \item \textbf{OCRBench}~\cite{Liu_2024OCRBench}: OCRBench is a comprehensive benchmark designed to evaluate the OCR capabilities of MLLMs. It comprises 1,000 question-answer pairs covering five specific components: text recognition, scene-text VQA, document-oriented VQA, key information extraction, and handwritten mathematical expression recognition.
    
    \item \textbf{ScienceQA}~\cite{SQA}: ScienceQA is a multimodal dataset consisting of multiple-choice science questions covering diverse topics (natural science, social science, language science). It features a rich set of lectures and explanations, supporting Chain-of-Thought (CoT) reasoning evaluation for elementary to high school-level science problems.
    
    \item \textbf{AI2D}~\cite{kembhavi2016diagramworthdozenimages}: AI2D (Allen Institute for Artificial Intelligence Diagrams) is a dataset focused on diagram understanding. It involves answering multiple-choice questions about scientific diagrams, requiring models to parse arrows, text labels, and the relationships between graphical elements.
    
    \item \textbf{ChartQA}~\cite{masry-etal-2022-chartqa}: ChartQA is designed to evaluate reasoning over charts. It includes both human-written and machine-generated questions concerning bar charts, line charts, and pie charts. The tasks involve data extraction and complex reasoning about trends and statistics depicted in the visualizations.

    \item \textbf{DocVQA}~\cite{mathew2021docvqadatasetvqadocument}: DocVQA is a document visual question answering benchmark that requires models to answer natural-language questions based on document images. Since the answers often depend on small text regions, table structures, and spatial layouts, this benchmark is particularly sensitive to whether pruning methods can preserve fine-grained textual and structural cues.

    \item \textbf{InfoVQA}~\cite{mathew2021infographicvqa}: InfoVQA focuses on question answering over infographic images, which usually contain dense text, icons, charts, and complex layouts. Compared with ordinary scene-centric VQA, InfoVQA places greater emphasis on locating relevant textual evidence and integrating it with visual and layout information, making it a challenging testbed for text-heavy visual token pruning.
\end{itemize}

\subsubsection{Comprehensive Multimodal Benchmarks}
\begin{itemize}
    \item \textbf{MME}~\cite{fu2025mmecomprehensiveevaluationbenchmark}: MME is a comprehensive evaluation suite for MLLMs that assesses both perception and cognition. It includes 14 subtasks (e.g., existence, count, position, color, OCR, commonsense reasoning). The evaluation relies on a strict instruction-following protocol (Yes/No answers) to avoid ambiguity in scoring.
    
    \item \textbf{MMBench}~\cite{liu2024mmbenchmultimodalmodelallaround}: MMBench is a robust evaluation pipeline designed to assess various abilities of MLLMs using a circular evaluation strategy (shuffling options) to mitigate the impact of option bias. It covers diverse ability dimensions such as coarse-grained perception, fine-grained perception, and logic reasoning.
    
    \item \textbf{MMBench\_cn}~\cite{liu2024mmbenchmultimodalmodelallaround}: This is the Chinese language version of the MMBench dataset, designed to evaluate the multimodal understanding and reasoning capabilities of models in a Chinese linguistic context.
    
    \item \textbf{MM-Vet}~\cite{yu2024mmvetevaluatinglargemultimodal}: MM-Vet evaluates MLLMs on diverse multimodal tasks including recognition, OCR, knowledge, language generation, spatial awareness, and math. It uses an LLM-based evaluation metric (e.g., GPT-4) to score open-ended model outputs against ground truths, aiming to better capture the nuances of model performance.
\end{itemize}

\subsubsection{Hallucination and Trustworthiness}
\begin{itemize}
    \item \textbf{POPE}~\cite{li-etal-2023-evaluating}: The Polling-based Object Probing Evaluation (POPE) dataset is designed to evaluate object hallucination in MLLMs. It employs a polling strategy with three sampling settings (random, popular, and adversarial) to verify whether a model accurately predicts the existence of objects in an image or hallucinates non-existent ones.
    
    \item \textbf{HallusionBench}~\cite{Guan_2024_CVPR}: Often referred to as HallBench, this is an advanced diagnostic suite for disentangling language hallucination and visual illusion. It constructs visual-question control pairs (using both original and manipulated images) to rigorously test whether models rely on visual facts or parametric knowledge bias, focusing on nuanced understanding and logical consistency.
\end{itemize}

\subsubsection{Video Understanding}
\begin{itemize}
    \item \textbf{MVBench}~\cite{MVBench}: MVBench is a comprehensive multi-modal video understanding benchmark. It introduces a static-to-dynamic method to generate temporal tasks from static image datasets. It covers 20 challenging temporal tasks across categories like action sequence, object interaction, and scene transition, requiring models to process dynamic visual content effectively.
    
    \item \textbf{LongVideoBench}~\cite{LongVideoBench}: LongVideoBench focuses on long-context video understanding. It contains videos with durations ranging significantly, up to one hour. The benchmark tests the model's ability to handle long-term temporal dependencies and interleaved video-language contexts, featuring tasks such as referring reasoning over extensive video content.
    
    \item \textbf{Video-MME}~\cite{fu2025videommefirstevercomprehensiveevaluation}: Video-MME is a full-spectrum, multi-modal evaluation benchmark for video analysis. It distinguishes itself by covering a wide range of video durations (short, medium, and long up to 60 minutes) and diverse domains (knowledge, film, sports, etc.). It also incorporates multi-modal inputs including video frames, subtitles, and audio to comprehensively assess the capabilities of MLLMs.
\end{itemize}

\subsection{Model Architectures}
In our experiments, we evaluate a diverse set of Multimodal Large Language Models (MLLMs) ranging from static image understanding to dynamic video processing. Additionally, we utilize specific vision encoders for  instruction relevance score computation. The details of these models are provided below.

\subsubsection{Large Vision-Language Models}
We select representative open-source MLLMs that are widely recognized in the community, covering both general-purpose and video-specific architectures.

\begin{itemize}
    \item \textbf{LLaVA-1.5}~\cite{liu2023llava}: LLaVA-1.5 is a seminal open-source MLLM that connects a pre-trained vision encoder (CLIP ViT-L/336px) with a large language model (Vicuna) using a simple two-layer MLP projection. It serves as a robust baseline for general visual question answering and instruction following tasks.
    
    \item \textbf{LLaVA-NeXT}~\cite{liu2024llavanext}: Also known as LLaVA-1.6, this model improves upon LLaVA-1.5 by incorporating stronger language backbones (such as Mistral and Vicuna-1.5) and increasing the input image resolution. It utilizes an ``AnyRes'' strategy to handle images with various aspect ratios and high resolutions, significantly enhancing performance on OCR and fine-grained visual tasks.
    
    \item \textbf{Qwen2.5-VL}~\cite{bai2025qwen25vltechnicalreport}: Qwen2.5-VL is one of the latest iteration in the Qwen-VL series, built upon the powerful Qwen2.5 language model. It features native support for dynamic resolution (NaViT), allowing it to process images of arbitrary aspect ratios and durations without padding. It demonstrates state-of-the-art performance across both image and video understanding benchmarks.

    \item \textbf{Qwen3-VL}~\cite{bai2025qwen3vltechnicalreport}: Qwen3-VL is a recent Qwen-family vision-language model with improved multimodal perception, reasoning, long-context understanding, and text-rich visual understanding. It introduces stronger visual-language alignment and architectural upgrades for spatial-temporal modeling and multi-level visual feature integration. We evaluate EADP on Qwen3-VL-8B with 1024 visual tokens as the full-token setting. To better examine pruning robustness in text-heavy scenarios, we additionally include DocVQA and InfoVQA on top of the common Qwen-series benchmarks.
    
    \item \textbf{LLaVA-Video}~\cite{zhang2025llavavideovideoinstructiontuning}: LLaVA-Video is a specialized MLLM designed to handle long-context video understanding. Unlike the standard LLaVA series which typically uses CLIP, LLaVA-Video adopts a stronger vision tower (SigLIP) and employs spatial-temporal pooling to efficiently process a large number of frames, making it highly effective for temporal reasoning tasks.
\end{itemize}

\subsubsection{Vision Encoders}
Vision encoders serve as the visual perception module for MLLMs. In this work, they also play a crucial role in our proposed methodology.

\begin{itemize}
    \item \textbf{CLIP}~\cite{radford2021clip}: The Contrastive Language-Image Pre-training (CLIP) model aligns visual and textual representations in a shared embedding space. It serves as the visual backbone for LLaVA-1.5 and LLaVA-NeXT. Its pre-trained weights provide the fundamental visual semantics required for cross-modal alignment.
    
    \item \textbf{SigLIP}~\cite{zhai2023sigmoidlosslanguageimage}: SigLIP (Sigmoid Loss for Language Image Pre-training) replaces the standard softmax loss in contrastive learning with a sigmoid loss, resulting in better performance and training efficiency. It serves as the vision tower for LLaVA-Video~\cite{zhang2025llavavideovideoinstructiontuning}. Notably, in our method, we utilize SigLIP beyond its role as a backbone; we leverage it to explicitly measure the semantic relevance between image tokens and text tokens in experiments on LLaVA-Video~\cite{zhang2025cdpruner}.
\end{itemize}

\subsection{Comparison Methods}

To rigorously evaluate the effectiveness of our proposed approach, we compare it against a diverse set of state-of-the-art visual token prune baselines.

\begin{itemize}
    \item \textbf{FastV}~\cite{chen2024fastv}: This method identifies the phenomenon of inefficient visual attention allocation in VLMs. It accelerates inference by directly discarding visual tokens that receive the lowest attention weights from the language model after the initial shallow layers.
    
    \item \textbf{PyramidDrop}~\cite{xing2025pyramiddropacceleratinglargevisionlanguage}: Building upon the observation that visual token redundancy increases with model depth, PyramidDrop introduces a hierarchical, stage-wise pruning strategy. It progressively drops a certain proportion of visual tokens at different depths of the LLM, preserving performance while reducing computational overhead.
    
    \item \textbf{SparseVLM}~\cite{zhang2024sparsevlm}: Instead of relying on global attention, SparseVLM employs a multi-stage pruning strategy guided by specific text tokens. It selects the text tokens most relevant to the visual input to act as ``raters,'' and utilizes their specific attention weights towards visual tokens to guide the dropping process, yielding more precise reduction.

    \item \textbf{LLaVA-Prumerge (and Prumerge+)}~\cite{shang2025prumerge}: This approach focuses on clustering rather than simple dropping. It first identifies important anchor tokens based on the attention scores within the vision encoder. Subsequently, it aggregates the remaining redundant tokens into their most similar anchors via clustering, thereby preserving essential visual information.
    
    \item \textbf{VisionZip}~\cite{yang2025visionzip}: Observing that visual attention is highly concentrated, VisionZip extracts a subset of dominant tokens based on vision encoder attention. It then applies clustering to the remaining tokens to capture contextual background information, ultimately combining both sets to maintain a holistic visual representation.
    
    \item \textbf{HiPrune}~\cite{liu2025hiprune}: A training-free and model-agnostic framework that exploits the inherent hierarchical attention structure within vision encoders. It systematically partitions tokens into three categories: object-centric ``anchor'' tokens selected from middle layers, spatial ``buffer'' tokens adjacent to anchors, and global ``register'' tokens selected from deep layers. This structured selection balances local details and global context.

    \item \textbf{DART}~\cite{wen2025dart}: Arguing that eliminating duplication is more critical than selecting based on importance, DART employs a greedy approach to find a highly diverse subset of tokens. It selects a set of pivot tokens and iteratively retains remaining tokens that exhibit the lowest similarity to the already selected ones.
    
    \item \textbf{DivPrune}~\cite{alvar2025divprune}: This method formulates the token pruning task as a Max-Min Diversity Problem (MMDP). It seeks to retain a subset of visual tokens that maximizes the minimum pairwise feature distance among them, ensuring that the selected tokens cover a broad semantic space.

    \item \textbf{TRIM}~\cite{song2025trim}: TRIM addresses the limitation of purely vision-based pruning by leveraging cross-modal alignment metrics. It computes the cosine similarity between the image tokens (from the vision encoder) and the text tokens (from the text encoder), subsequently pruning the visual tokens that exhibit low relevance to the text prompt.
    
    \item \textbf{CDPruner}~\cite{zhang2025cdpruner}: This approach maximizes the \textit{conditional diversity} of the selected visual tokens. It reformulates the pruning process using a Determinantal Point Process (DPP). By simultaneously modeling the pairwise feature similarity between visual tokens and their relevance to the specific user instruction, CDPruner ensures that the retained subset is both highly representative of the image and strictly aligned with the query.
\end{itemize}

\section{More Results}
\label{sec:supp_more}

\subsection{Detailed Performance on MVBench}

To supplement the video evaluation results presented in the main text, we provide a comprehensive, subtask-level performance breakdown on MVBench. 
As shown in \cref{tab:mvbench_details}, EADP consistently outperforms existing pruning baselines across all three token reduction settings on LLaVA-Video-7B. Notably, even under the extreme compression regime where only $64 \times 16$ tokens are retained, EADP maintains a highly competitive average score of 52.6, surpassing strong baselines such as DivPrune (52.1) and CDPruner (50.2). 

\noindent A closer inspection of the 20 individual subtasks reveals EADP's distinct advantages in scenarios demanding precise spatial-temporal alignment. For instance, on tasks requiring detailed frame-level discrimination—such as \textit{action localization}, \textit{fine grained action}, and \textit{object interaction}—EADP exhibits noticeable and consistent margins over competing methods across various budgets. This subtask-level superiority further demonstrates that EADP effectively identifies and preserves the most informative keyframes and spatial patches critical for complex video reasoning, while safely discarding massive amounts of redundant temporal background.

\begin{table}[t]
    \centering
    \caption{\textbf{Performance comparison details on MVBench.} Avg. denotes the average over 20 subtasks.}
    \label{tab:mvbench_details}
    \resizebox{\linewidth}{!}{
    \small
    \begin{tabular}{lrrrrrrrrrrrrrrrrrrrrr}  
    \toprule
    \textbf{Method} & \rotatebox{90}{\textbf{action antonym}} & \rotatebox{90}{\textbf{action count}} & \rotatebox{90}{\textbf{action localization}} & \rotatebox{90}{\textbf{action prediction}} & \rotatebox{90}{\textbf{action sequence}} & \rotatebox{90}{\textbf{character order}} & \rotatebox{90}{\textbf{counterfactual inference}} & \rotatebox{90}{\textbf{egocentric navigation}} & \rotatebox{90}{\textbf{episodic reasoning}} & \rotatebox{90}{\textbf{fine grained action}} & \rotatebox{90}{\textbf{fine grained pose}} & \rotatebox{90}{\textbf{moving attribute}} & \rotatebox{90}{\textbf{moving count}} & \rotatebox{90}{\textbf{moving direction}} & \rotatebox{90}{\textbf{object existence}} & \rotatebox{90}{\textbf{object interaction}} & \rotatebox{90}{\textbf{object shuffle}} & \rotatebox{90}{\textbf{scene transition}} & \rotatebox{90}{\textbf{state change}} & \rotatebox{90}{\textbf{unexpected action}} & \textbf{Avg.} \\
    \midrule
    \rowcolor{lightgray!75}\multicolumn{22}{c}{\textit{Upper Bound, All 64 $\times$ 169 Tokens} ($\mathbf{100\%}$)} \\
    \rowcolor{lightgray!25} LLaVA-Video-7B & 77.0 & 57.5 & 61.5 & 63.5 & 72.0 & 74.5 & 48.5 & 30.0 & 53.0 & 49.0 & 56.0 & 71.5 & 44.5 & 36.0 & 61.0 & 84.0 & 41.0 & 92.5 & 54.0 & 81.0 & 60.4 \\
    \rowcolor{lightgray!75}\multicolumn{22}{c}{\textit{Retain 64 $\times$ 64 Tokens} \textcolor{Green} {($\downarrow\mathbf{62.1\%}$)}} \\
    DivPrune{\small\texttt{(CVPR25)}} & 67.5 & 38.5 & 45.5 & 55.5 & 68.0 & 73.0 & 37.5 & 31.0 & 52.5 & 46.0 & 48.5 & 64.5 & 45.0 & 35.5 & 49.5 & 80.5 & 41.5 & 91.0 & 53.5 & 79.5 & 55.2 \\
    CDPruner{\small\texttt{(NIPS25)}} & 66.0 & 38.5 & 46.5 & 54.5 & 65.5 & 70.5 & 37.5 & 32.0 & 52.0 & 47.0 & 48.0 & 58.5 & 42.5 & 35.5 & 49.0 & 74.5 & 40.5 & 89.5 & 54.0 & 80.5 & 54.1 \\
    HiPrune{\small\texttt{(AAAI26)}} & 68.0 & 40.0 & 44.5 & 57.0 & 61.5 & 61.5 & 44.0 & 35.0 & 53.0 & 45.5 & 33.0 & 69.5 & 35.5 & 38.0 & 51.0 & 77.0 & 41.5 & 91.0 & 52.5 & 80.0 & 54.0 \\
    \textbf{EADP}{\small\texttt{(Ours)}} & \textbf{66.5} & \textbf{39.0} & \textbf{52.0} & \textbf{57.0} & \textbf{66.5} & \textbf{71.0} & \textbf{43.0} & \textbf{32.5} & \textbf{54.0} & \textbf{48.0} & \textbf{50.0} & \textbf{60.5} & \textbf{40.5} & \textbf{36.0} & \textbf{49.5} & \textbf{79.0} & \textbf{43.5} & \textbf{88.5} & \textbf{55.0} & \textbf{81.0} & \textbf{55.7} \\
    \rowcolor{lightgray!75}\multicolumn{22}{c}{\textit{Retain 64 $\times$ 32 Tokens} \textcolor{Green} {($\downarrow\mathbf{81.1\%}$)}} \\
    DivPrune{\small\texttt{(CVPR25)}} & 68.5 & 39.5 & 42.0 & 55.5 & 61.0 & 71.5 & 35.5 & 29.5 & 54.0 & 47.0 & 45.0 & 59.5 & 47.5 & 33.5 & 49.5 & 76.5 & 38.5 & 90.0 & 52.5 & 77.5 & 53.7 \\
    CDPruner{\small\texttt{(NIPS25)}} & 71.0 & 38.5 & 41.5 & 54.0 & 63.5 & 68.5 & 37.5 & 31.0 & 51.0 & 46.0 & 45.5 & 51.5 & 45.5 & 33.5 & 50.0 & 70.5 & 42.0 & 89.0 & 54.0 & 79.0 & 53.2 \\
    HiPrune{\small\texttt{(AAAI26)}} & 70.0 & 42.5 & 39.0 & 49.5 & 59.5 & 50.0 & 40.5 & 33.5 & 53.0 & 44.5 & 25.5 & 64.0 & 41.0 & 36.0 & 46.0 & 73.5 & 41.0 & 91.0 & 46.0 & 75.5 & 51.1 \\
    \textbf{EADP}{\small\texttt{(Ours)}} & \textbf{72.5} & \textbf{41.0} & \textbf{44.0} & \textbf{56.5} & \textbf{65.5} & \textbf{68.0} & \textbf{44.0} & \textbf{33.0} & \textbf{52.5} & \textbf{45.5} & \textbf{49.0} & \textbf{56.0} & \textbf{43.5} & \textbf{34.0} & \textbf{48.0} & \textbf{76.0} & \textbf{42.5} & \textbf{86.0} & \textbf{54.5} & \textbf{77.0} & \textbf{54.5} \\
    \rowcolor{lightgray!75}\multicolumn{22}{c}{\textit{Retain 64 $\times$ 16 Tokens} \textcolor{Green}{($\downarrow\mathbf{90.5\%}$)}} \\
    DivPrune{\small\texttt{(CVPR25)}} & 73.5 & 41.5 & 40.0 & 52.5 & 60.0 & 64.5 & 35.0 & 29.5 & 54.0 & 43.5 & 42.5 & 58.0 & 47.5 & 31.5 & 49.5 & 67.5 & 39.5 & 87.5 & 48.5 & 76.5 & 52.1 \\
    CDPruner{\small\texttt{(NIPS25)}} & 73.0 & 38.5 & 39.5 & 50.0 & 57.0 & 59.5 & 41.0 & 30.0 & 49.5 & 40.0 & 43.5 & 49.0 & 42.5 & 31.0 & 46.5 & 66.0 & 42.0 & 86.5 & 44.5 & 74.5 & 50.2 \\
    HiPrune{\small\texttt{(AAAI26)}} & 73.5 & 44.5 & 36.0 & 47.0 & 50.0 & 46.0 & 35.5 & 33.0 & 49.5 & 42.0 & 22.0 & 65.5 & 45.0 & 34.0 & 50.0 & 61.5 & 37.5 & 89.5 & 40.0 & 73.0 & 48.8 \\
    \textbf{EADP}{\small\texttt{(Ours)}} & \textbf{71.5} & \textbf{40.5} & \textbf{43.0} & \textbf{52.0} & \textbf{63.0} & \textbf{55.0} & \textbf{38.5} & \textbf{33.5} & \textbf{51.5} & \textbf{44.5} & \textbf{43.0} & \textbf{58.5} & \textbf{46.0} & \textbf{35.5} & \textbf{50.0} & \textbf{69.5} & \textbf{41.0} & \textbf{86.5} & \textbf{51.5} & \textbf{78.0} & \textbf{52.6} \\
    \bottomrule
    \end{tabular}
    }
\end{table}

\begin{table}[t]
    \centering
    \caption{\textbf{Performance comparison on LLaVA-1.5-13B.} Avg. denotes the average performance across 9 benchmarks.}
    \label{tab:llava-1.5-13b}
    \resizebox{\linewidth}{!}{
      \begin{tabular}{l|ccccccccc|c}
      \toprule
      \textbf{Method} & \textbf{$\text{VQA}^\text{V2}$} & \textbf{GQA} & \textbf{$\text{SQA}^\text{IMG}$} & \textbf{$\text{VQA}^\text{Text}$} & \textbf{POPE} & \textbf{MME} & \textbf{$\text{MMB}^\text{EN}$} & \textbf{$\text{MMB}^\text{CN}$} & \textbf{MMVet} & \textbf{Acc.} \\
      \rowcolor{lightgray!75}\multicolumn{11}{c}{\textit{Upper Bound, All 576 Tokens} ($\mathbf{100\%}$)} \\
      \rowcolor{lightgray!25} LLaVA-1.5-13B & 80.0 & 63.3 & 72.8 & 61.2 & 86.0 & 1531.2 & 68.5 & 63.5 & 36.2 & 67.6 \\
      \rowcolor{lightgray!75}\multicolumn{11}{c}{\textit{Retain 128 Tokens} \textcolor{Green}{($\downarrow\mathbf{77.8\%}$)}} \\
      FastV{\small\texttt{(ECCV24)}}        & 73.6 & 57.2 & 72.1 & 55.9 & 72.9 & 1421.5 & 64.6 & 59.9 & 33.4 &  62.3\\
      PDrop{\small\texttt{(CVPR25)}}        & 77.1 & 60.5 &  71.5 & 57.2 & 81.8 & 1476.1 & 66.0 & 60.6 & 31.8 &  64.5\\
      SparseVLM{\small\texttt{(ICML25)}}    & 76.2 & 58.9 &  72.1 & 56.3 & 83.1 & 1454.2 & 66.1 & 60.8 & 34.9 &  64.6\\
      PruMerge+{\small\texttt{(2024.05)}}   & 75.7 & 57.6 &  71.9 & 54.4 & 81.5 & 1424.9 & 64.5 & 58.9 & 34.7 &  63.4\\
      TRIM{\small\texttt{(COLING25)}}       & 74.5 & 58.5 &  70.6 & 53.0 & 85.2 & 1417.2 & 65.3 & 56.7 & 35.6 &  63.4\\
      VisionZip{\small\texttt{(CVPR25)}}    & 76.2 & 57.1 &  71.8 & 56.4 & 81.5 & 1422.9 & 65.8 & 60.4 & 35.5 &  64.0\\
      DART{\small\texttt{(EMNLP25)}}        & 74.8 & 57.0 &  72.4 & 52.5 & 78.3 & 1387.8 & 63.6 & 60.1 & 32.9 &  62.3\\
      DivPrune{\small\texttt{(CVPR25)}}     & 76.3 & 58.6 &  71.4 & 57.2 & 83.6 & 1444.3 & 64.5 & 58.9 & 33.1 &  64.0\\
      CDPruner{\small\texttt{(NIPS25)}}     & 77.2 & 59.2 &  71.8 & 57.1 & 85.3 & 1465.0 & 65.9 & 60.9 & 36.2 &  65.2\\
      \textbf{EADP}{\small\texttt{(Ours)}}  & \textbf{77.9} & \textbf{60.0} & \textbf{72.8} & \textbf{57.9} & \textbf{86.5} & \textbf{1468.1} & \textbf{66.5} & \textbf{61.3} & \textbf{37.9} & \textbf{66.0} \\
      \rowcolor{lightgray!75}\multicolumn{11}{c}{\textit{Retain 64 Tokens} \textcolor{Green}{($\downarrow\mathbf{88.9\%}$)}} \\
      FastV{\small\texttt{(ECCV24)}}        & 66.1 & 50.3 &  70.7 & 52.1 & 57.4 & 1232.1 & 58.7 & 57.3 & 27.5 & 55.7 \\
      PDrop{\small\texttt{(CVPR25)}}        & 68.6 & 53.7 &  70.1 & 53.9 & 68.1 & 1239.5 & 62.4 & 57.8 & 24.4 & 57.9 \\
      SparseVLM{\small\texttt{(ICML25)}}    & 71.7 & 55.1 &  70.9 & 55.8 & 76.6 & 1334.9 & 64.5 & 59.2 & 31.4 & 61.3 \\
      PruMerge+{\small\texttt{(2024.05)}}   & 70.3 & 54.8 &  71.3 & 53.8 & 76.3 & 1364.1 & 63.7 & 57.1 & 31.9 & 60.8 \\
      TRIM{\small\texttt{(COLING25)}}       & 71.5 & 55.2 &  69.8 & 51.8 & 85.9 & 1389.6 & 64.2 & 53.8 & 29.6 & 61.3 \\
      VisionZip{\small\texttt{(CVPR25)}}    & 72.9 & 57.1 &  71.0 & 55.7 & 78.9 & 1357.5 & 64.1 & 58.9 & 32.6 & 62.1 \\
      DART{\small\texttt{(EMNLP25)}}        & 72.3 & 56.3 &  71.7 & 56.6 & 73.3 & 1362.4 & 62.5 & 59.4 & 31.6 & 61.3 \\
      DivPrune{\small\texttt{(CVPR25)}}     & 74.8 & 57.2 &  70.5 & 56.8 & 82.1 & 1423.7 & 63.2 & 58.3 & 30.2 & 62.7 \\
      CDPruner{\small\texttt{(NIPS25)}}     & 76.2 & 58.9 &  71.4 & 56.9 & 84.9 & 1453.0 & 64.6 & 59.5 & 34.5 & 64.4 \\
      \textbf{EADP}{\small\texttt{(Ours)}}  & \textbf{77.1} & \textbf{59.6} & \textbf{72.5} & \textbf{57.2} & \textbf{86.6} & \textbf{1461.4} & \textbf{65.3} & \textbf{60.7} & \textbf{33.7} & \textbf{65.1} \\
      \rowcolor{lightgray!75}\multicolumn{11}{c}{\textit{Retain 32 Tokens} \textcolor{Green}{($\downarrow\mathbf{94.4\%}$)}} \\
      PruMerge+{\small\texttt{(2024.05)}}   & 64.3 & 53.2 &  70.0 & 51.2 & 68.9 & 1307.3 & 60.5 & 53.0 & 27.1 & 57.1 \\
      TRIM{\small\texttt{(COLING25)}}       & 67.9 & 53.5 &  68.3 & 50.3 & 83.4 & 1345.2 & 62.6 & 47.9 & 26.8 & 58.7 \\
      VisionZip{\small\texttt{(CVPR25)}}    & 69.1 & 53.1 &  70.4 & 54.4 & 69.7 & 1299.2 & 61.7 & 53.2 & 28.7 &  58.4\\
      DART{\small\texttt{(EMNLP25)}}        & 68.5 & 53.4 &  71.1 & 53.9 & 68.2 & 1321.7 & 60.8 & 56.4 & 29.1 & 58.6 \\
      DivPrune{\small\texttt{(CVPR25)}}     & 71.7 & 55.8 &  69.8 & 54.1 & 78.9 & 1389.2 & 61.8 & 56.0 & 28.2 & 60.6 \\
      CDPruner{\small\texttt{(NIPS25)}}     & 74.7 & 57.9 &  71.5 & 54.7 & 84.5 & 1401.7 & 63.4 & 56.6 & 30.9 & 62.7 \\
      \textbf{EADP}{\small\texttt{(Ours)}}  & \textbf{75.8} & \textbf{58.8} & \textbf{72.9} & \textbf{55.2} & \textbf{86.1} & \textbf{1409.2} & \textbf{64.2} & \textbf{57.3} & \textbf{31.4} & \textbf{63.6} \\
      \bottomrule
      \end{tabular}
    }
\end{table}

\subsection{Performance on More Architectures and Scales}

To further demonstrate the scalability of our method, we extend the performance evaluation across a wider spectrum of model capacities, including scaling up to 13B backbones (LLaVA-1.5-13B and LLaVA-NeXT-13B) and scaling down to a highly compact footprint (Qwen2.5-VL-3B).

\begin{table}[t]
    \centering
    \caption{\textbf{Performance comparison on LLaVA-NEXT-13B.} Avg. denotes the average performance across 9 benchmarks.}
    \label{tab:llava-next-13b}
    \resizebox{\linewidth}{!}{
      \begin{tabular}{l|ccccccccc|c}
      \toprule
      \textbf{Method} & \textbf{$\text{VQA}^\text{V2}$} & \textbf{GQA} & \textbf{$\text{SQA}^\text{IMG}$} & \textbf{$\text{VQA}^\text{Text}$} & \textbf{POPE} & \textbf{MME} & \textbf{$\text{MMB}^\text{EN}$} & \textbf{$\text{MMB}^\text{CN}$} & \textbf{MMVet} & \textbf{Acc.} \\
      \rowcolor{lightgray!75}\multicolumn{11}{c}{\textit{Upper Bound, All 2880 Tokens} ($\mathbf{100\%}$)} \\
      \rowcolor{lightgray!25} LLaVA-NeXT-13B & 82.3 & 64.4 & 73.1 & 63.2 & 85.3 & 1539.5 & 68.5 & 61.2 & 45.0 & 68.9 \\
      \rowcolor{lightgray!75}\multicolumn{11}{c}{\textit{Retain 640 Tokens} \textcolor{Green}{($\downarrow\mathbf{77.8\%}$)}} \\
      FastV{\small\texttt{(ECCV24)}}        & 77.1 & 60.6 & 70.1 & 57.9 & 81.3 & 1478.3 & 64.3 & 58.5 & 39.5 & 64.8 \\
      PDrop{\small\texttt{(CVPR25)}}        & 78.3 & 61.3 & 69.8 & 59.3 & 83.5 & 1522.9 & 65.4 & 60.4 & 38.1 &  65.8\\
      SparseVLM{\small\texttt{(ICML25)}}    & 77.5 & 61.4 & 71.2 & 58.7 & 84.7 & 1543.3 & 67.2 & 62.5 & 39.6 & 66.7 \\
      PruMerge+{\small\texttt{(2024.05)}}   & 76.2 & 61.9 & 68.5 & 54.5 & 82.9 & 1476.2 & 66.9 & 60.8 & 37.7 & 64.8 \\
      TRIM{\small\texttt{(COLING25)}}       & 78.1 & 61.3 & 69.2 & 55.3 & 84.3 & 1531.8 & 67.3 & 61.3 & 40.1 & 65.9 \\
      VisionZip{\small\texttt{(CVPR25)}}    & 78.3 & 60.8 & 68.7 & 58.7 & 84.7 & 1528.1 & 66.6 & 60.4 & 42.4 & 66.3 \\
      DART{\small\texttt{(EMNLP25)}}        & 77.9 & 61.6 & 69.5 & 59.2 & 85.3 & 1517.9 & 66.3 & 61.1 & 41.1 & 66.4 \\
      DivPrune{\small\texttt{(CVPR25)}}     & 79.2 & 62.8 & 71.1 & 57.8 & 85.5 & 1502.4 & 67.1 & 61.7 & 40.5 & 66.8 \\
      CDPruner{\small\texttt{(NIPS25)}}     & 80.1 & 63.4 & 71.9 & 59.8 & 86.8 & 1540.7 & 68.3 & 62.1 & 40.3 & 67.7 \\
      \textbf{EADP}{\small\texttt{(Ours)}}  & \textbf{80.6} & \textbf{64.0} & \textbf{72.7} & \textbf{60.4} & \textbf{87.4} & \textbf{1580.9} & \textbf{69.2} & \textbf{62.9} & \textbf{41.3} & \textbf{68.6} \\
      \rowcolor{lightgray!75}\multicolumn{11}{c}{\textit{Retain 320 Tokens} \textcolor{Green}{($\downarrow\mathbf{88.9\%}$)}} \\
      FastV{\small\texttt{(ECCV24)}}        & 67.1 & 53.9 & 69.1 & 54.8 & 67.2 & 1288.1 & 59.6 & 55.7 & 32.8 & 58.3 \\
      PDrop{\small\texttt{(CVPR25)}}        & 73.2 & 55.1 & 69.4 & 55.3 & 75.8 & 1375.4 & 60.8 & 57.1 & 30.1 & 60.6 \\
      SparseVLM{\small\texttt{(ICML25)}}    & 74.4 & 60.5 & 70.1 & 57.0 & 81.3 & 1489.2 & 64.0 & 62.0 & 36.7 & 64.5 \\
      PruMerge+{\small\texttt{(2024.05)}}   & 74.7 & 59.8 & 68.2 & 53.1 & 80.6 & 1445.1 & 63.2 & 59.3 & 34.1 & 62.8 \\
      TRIM{\small\texttt{(COLING25)}}       & 75.0 & 59.5 & 68.8 & 50.6 & 83.9 & 1462.3 & 65.1 & 56.8 & 33.9 & 63.0 \\
      VisionZip{\small\texttt{(CVPR25)}}    & 74.8 & 59.1 & 68.3 & 57.1 & 83.5 & 1501.3 & 63.9 & 59.2 & 39.7 & 64.5 \\
      DART{\small\texttt{(EMNLP25)}}        & 75.1 & 60.2 & 68.6 & 57.2 & 84.2 & 1489.9 & 64.2 & 60.4 & 39.6 & 64.9 \\
      DivPrune{\small\texttt{(CVPR25)}}     & 77.2 & 61.9 & 70.3 & 55.9 & 84.5 & 1477.5 & 64.6 & 60.5 & 37.8 & 65.2 \\
      CDPruner{\small\texttt{(NIPS25)}}     & 78.3 & 62.9 & 70.8 & 57.3 & 86.2 & 1511.4 & 65.9 & 61.0 & 40.6 & 66.5 \\
      \textbf{EADP}{\small\texttt{(Ours)}}  & \textbf{79.3} & \textbf{63.6} & \textbf{71.6} & \textbf{58.1} & \textbf{87.5} & \textbf{1559.9} & \textbf{66.3} & \textbf{61.4} & \textbf{41.2} & \textbf{67.4} \\
      \rowcolor{lightgray!75}\multicolumn{11}{c}{\textit{Retain 160 Tokens} \textcolor{Green}{($\downarrow\mathbf{94.4\%}$)}} \\
      PruMerge+{\small\texttt{(2024.05)}}   & 72.8 & 55.6 & 66.4 & 50.2 & 74.8 & 1367.2 & 61.4 & 57.3 & 32.2 & 59.9 \\
      TRIM{\small\texttt{(COLING25)}}       & 73.6 & 55.9 & 67.0 & 47.8 & 82.1 & 1410.1 & 63.9 & 52.9 & 29.4 &  60.3 \\
      VisionZip{\small\texttt{(CVPR25)}}    & 73.9 & 58.6 & 67.9 & 54.3 & 82.8 & 1432.7 & 63.6 & 58.7 & 34.2 & 62.8 \\
      DART{\small\texttt{(EMNLP25)}}        & 74.3 & 59.4 & 69.1 & 54.1 & 79.7 & 1467.1 & 63.7 & 58.8 & 35.2 &  63.1 \\
      DivPrune{\small\texttt{(CVPR25)}}     & 75.1 & 60.2 & 69.9 & 54.0 & 81.8 & 1412.0 & 64.3 & 59.1 & 35.8 & 63.4 \\
      CDPruner{\small\texttt{(NIPS25)}}     & 76.9 & 61.7 & 71.3 & 55.1 & 85.4 & 1487.9 & 65.6 & 59.7 & 37.7 &  65.3 \\
      \textbf{EADP}{\small\texttt{(Ours)}}  & \textbf{77.8} & \textbf{62.4} & \textbf{72.1} & \textbf{55.6} & \textbf{87.2} & \textbf{1524.2} & \textbf{66.3} & \textbf{60.9} & \textbf{36.9} & \textbf{66.2} \\
      \bottomrule
      \end{tabular}
    }
\end{table}

\noindent \textbf{Scalability on 13B LLM Backbones.}
As shown in \cref{tab:llava-1.5-13b} and \cref{tab:llava-next-13b}, scaling the language backbone from 7B to 13B yields consistent overall improvements, and EADP seamlessly translates its superiority to these larger capacities. On the standard-resolution LLaVA-1.5-13B (\cref{tab:llava-1.5-13b}), EADP consistently establishes new state-of-the-art results across all budgets. More impressively, on the high-resolution LLaVA-NeXT-13B (\cref{tab:llava-next-13b}), EADP achieves near-lossless compression at the 640-token budget (averaging 68.6, remarkably close to the 68.9 upper bound). Even under the extreme reduction where $94.4\%$ of visual tokens are discarded, EADP maintains a highly competitive score of 66.2. This confirms that our submodular modeling effectively scales with both model capacity and input resolution, consistently preserving semantic integrity.

\noindent \textbf{Robustness on Compact Model Capacities.}
Complementing the Qwen2.5-VL-7B evaluation in the main text, \cref{tab:qwen2.5-vl-3b} assesses EADP on the significantly smaller Qwen2.5-VL-3B. Visual token pruning for compact LLMs is inherently challenging, as their limited parameter count makes them more vulnerable to context loss. While EADP delivers highly competitive performance at moderate compression (e.g., retaining 512 tokens), its robustness shines brightest under aggressive reduction regimes. At the 128-token budget, EADP achieves a leading average score of 62.1, outperforming strong baselines like DivPrune (61.1) and HiPrune (56.8). This underscores EADP's exceptional capability to extract strictly vital visual cues, successfully guiding even lightweight VLMs to perform accurate reasoning when representational capacity is severely constrained.

\begin{table}[!t]
  \centering
  \caption{\textbf{Performance comparison on Qwen2.5-VL-3B.} Avg. denotes the average performance across 8 benchmarks.}
  \label{tab:qwen2.5-vl-3b}
  \resizebox{\linewidth}{!}{
    \begin{tabular}{l|cccccccc|c}
    \toprule
    \textbf{Method} & \textbf{TextVQA} & \textbf{ChartQA} & \textbf{AI2D} & \textbf{OCRBench} & \textbf{HallBench} & \textbf{MME} & \textbf{MMB-EN} & \textbf{MMB-CN} & \textbf{Avg.} \\
    \rowcolor{lightgray!75}\multicolumn{10}{c}{\textit{Upper Bound, All 1296 Tokens} ($\mathbf{100\%}$)} \\
    \rowcolor{lightgray!25} Qwen2.5-VL-3B & 78.8 & 83.1 & 81.3 & 791 & 36.1 & 2192.3 & 77.9 & 77.8 & 78.0 \\
    \rowcolor{lightgray!75}\multicolumn{10}{c}{\textit{Retain 512 Tokens} \textcolor{Green}{($\downarrow\mathbf{60.5\%}$)}} \\
    DivPrune{\small\texttt{(CVPR25)}} & 72.3 & 73.4 & 76.4 & 676 & 33.5 & 2034.2 & 74.8 & 74.1 & 71.7 \\
    CDPruner{\small\texttt{(NIPS25)}} & 52.2 & 59.2 & 71.1 & 519 & 29.5 & 1912.8 & 72.9 & 71.0 & 62.9 \\
    HiPrune{\small\texttt{(AAAI26)}} & 69.1 & 70.0 & 76.6 & 627 & 33.8 & 1951.4 & 74.7 & 73.3 & 69.7 \\
    \textbf{EADP}{\small\texttt{(Ours)}} & \textbf{70.7} & \textbf{71.2} & \textbf{76.3} & \textbf{667} & \textbf{32.1} & \textbf{1967.2} & \textbf{74.9} & \textbf{73.1} & \textbf{70.4} \\
    \rowcolor{lightgray!75}\multicolumn{10}{c}{\textit{Retain 256 Tokens} \textcolor{Green}{($\downarrow\mathbf{80.2\%}$)}} \\
    DivPrune{\small\texttt{(CVPR25)}} & 66.5 & 62.2 & 75.5 & 576 & 29.5 & 1934.6 & 73.3 & 72.8 & 66.8 \\
    CDPruner{\small\texttt{(NIPS25)}} & 41.4 & 47.3 & 68.1 & 381 & 25.4 & 1735.2 & 68.6 & 67.6 & 55.4 \\
    HiPrune{\small\texttt{(AAAI26)}} & 57.8 & 54.2 & 73.8 & 512 & 27.5 & 1956.9 & 73.0 & 72.8 & 63.5 \\
    \textbf{EADP}{\small\texttt{(Ours)}} & \textbf{66.3} & \textbf{62.5} & \textbf{74.9} & \textbf{600} & \textbf{29.5} & \textbf{1978.9} & \textbf{73.0} & \textbf{73.2} & \textbf{67.3} \\
    \rowcolor{lightgray!75}\multicolumn{10}{c}{\textit{Retain 128 Tokens} \textcolor{Green}{($\downarrow\mathbf{90.1\%}$)}} \\
    DivPrune{\small\texttt{(CVPR25)}} & 59.3 & 49.3 & 72.2 & 448 & 28.9 & 1880.2 & 70.7 & 69.9 & 61.1 \\
    CDPruner{\small\texttt{(NIPS25)}} & 31.2 & 35.8 & 66.2 & 266 & 22.8 & 1631.5 & 64.8 & 61.5 & 48.8 \\
    HiPrune{\small\texttt{(AAAI26)}} & 44.5 & 39.6 & 69.4 & 425 & 25.1 & 1829.2 & 70.8 & 71.0 & 56.8 \\
    \textbf{EADP}{\small\texttt{(Ours)}} & \textbf{58.9} & \textbf{48.1} & \textbf{72.9} & \textbf{483} & \textbf{28.6} & \textbf{1939.8} & \textbf{71.3} & \textbf{71.4} & \textbf{62.1} \\
    \bottomrule
    \end{tabular}
  }
\end{table}

\subsection{More Efficiency Analysis}

To complement the efficiency analysis of LLaVA-1.5-7B presented in the main text, we further evaluate EADP on two additional models: LLaVA-NeXT-7B (high-resolution setting) and Qwen2.5-VL-3B (advanced architecture).

\noindent \textbf{Efficiency on High-Resolution LLaVA-NeXT-7B.}
As shown in \cref{tab:efficiency-llava-1.6}, the efficiency gains observed in the main text seamlessly transfer to the high-resolution regime. LLaVA-NeXT-7B processes a significantly larger computational footprint. Under a conservative 128-token budget, EADP substantially reduces the computational cost to 5709.6G FLOPs while delivering a $2.9\times$ speedup in prefill time and a $2.5\times$ speedup in overall latency. At the extreme 32-token compression, EADP accelerates the end-to-end latency by $4.2\times$. Consistent with our previous observations, EADP achieves these massive speedups with only a marginal computational overhead compared to other approaches, which is a highly worthwhile trade-off for its superior performance. 

\noindent \textbf{Generalization to Qwen2.5-VL-3B.}
To verify that our method is not strictly tailored to the LLaVA family, we further conduct evaluation on Qwen2.5-VL-3B. As detailed in \cref{tab:efficiency-qwen2.5-vl}, EADP demonstrates exceptional prefill acceleration, peaking at a $5.9\times$ speedup under 128-token budget. 
More importantly, the end-to-end latency metrics on Qwen2.5-VL reveal a critical vulnerability in existing methods: while baselines like HiPrune suffer from severe latency degradation, EADP remains highly stable and robust. This suggests that the pruning mechanisms of certain baselines may conflict with Qwen2.5-VL's underlying generation or attention implementation, causing severe overheads during the decoding phase. Additionally, this phenomenon is strongly tied to the compact 3B model scale, where the computational cost of their pruning operations easily outweighs the time saved within the lightweight language backbone. In contrast, EADP consistently delivers positive latency acceleration across all token budgets, proving its superior compatibility and efficiency across diverse architectures.

\noindent \textbf{Wall-clock time analysis.} 
\cref{tab:wall-time-clock} reports the runtime breakdown of EADP on Qwen3-VL-8B. The proposed scoring and refinement steps are lightweight: global guidance, dense guidance, score fusion, and smoothing \& polarization together require only about 1.8 ms across all token budgets. The dominant overhead comes from facility location selection, whose cost increases from 37.28 ms to 143.73 ms when the retained token budget grows from 128 to 512, due to the larger number of greedy marginal-gain updates.
After pruning, the LLM prefill time is reduced from 320.51 ms to 77.60, 122.63, and 178.97 ms under 128, 256, and 512 retained tokens, respectively.
Counting both pruning overhead and pruned prefill, the 128- and 256-token settings still provide clear acceleration, while the 512-token setting remains roughly comparable to the unpruned baseline.
These results indicate that EADP is efficient under practical token budgets, and that facility location selection is the main component to optimize further.

\begin{table}[!t]
  \noindent 
  \begin{minipage}[t]{0.485\textwidth}
    \centering
    \captionof{table}{Efficiency analysis on LLaVA-NEXT-7B.}

    \resizebox{\linewidth}{!}{%
    \begin{tabular}{l|ccc}
    \toprule
    \textbf{Method} & \textbf{Prefill(ms)} & \textbf{Latency(ms)} & \textbf{FLOPs(G)} \\
    \rowcolor{lightgray!75}\multicolumn{4}{c}{\textit{Upper Bound, All 576 Tokens} ($\mathbf{100\%}$)} \\
    \rowcolor{lightgray!25} LLaVA-1.6-7B & 743.7 & 819.7 & 12655.2 \\
    \rowcolor{lightgray!75}\multicolumn{4}{c}{\textit{Retain 128 Tokens} \textcolor{Green}{($\downarrow\mathbf{77.8\%}$)}} \\
    DivPrune{\small\texttt{(CVPR25)}} & 279.1 ($\times$2.7) & 347.0 ($\times$2.4) & 5693.6 ($\times$2.2) \\
    CDPruner{\small\texttt{(NIPS25)}} & 277.8 ($\times$2.7) & 340.4 ($\times$2.4) & 5709.4 ($\times$2.2) \\
    HiPrune{\small\texttt{(AAAI26)}} & 234.8 ($\times$3.2) & 299.9 ($\times$2.7) & 5693.6 ($\times$2.2) \\
    \textbf{EADP}{\small\texttt{(Ours)}} & \textbf{256.6 ($\times$2.9)} & \textbf{323.7 ($\times$2.5)} & \textbf{5709.6 ($\times$2.2)} \\
    \rowcolor{lightgray!75}\multicolumn{4}{c}{\textit{Retain 64 Tokens} \textcolor{Green}{($\downarrow\mathbf{88.9\%}$)}} \\
    DivPrune{\small\texttt{(CVPR25)}} & 198.2 ($\times$3.8) & 262.3 ($\times$3.1) & 3579.4 ($\times$3.5) \\
    CDPruner{\small\texttt{(NIPS25)}} & 196.6 ($\times$3.8) & 255.2 ($\times$3.2) & 3595.1 ($\times$3.5) \\
    HiPrune{\small\texttt{(AAAI26)}} & 176.5 ($\times$4.2) & 245.2 ($\times$3.3) & 3579.4 ($\times$3.5) \\
    \textbf{EADP}{\small\texttt{(Ours)}} & \textbf{188.2 ($\times$4.0)} & \textbf{257.2 ($\times$3.2)} & \textbf{3595.3 ($\times$3.5)} \\
    \rowcolor{lightgray!75}\multicolumn{4}{c}{\textit{Retain 32 Tokens} \textcolor{Green}{($\downarrow\mathbf{94.4\%}$)}} \\
    DivPrune{\small\texttt{(CVPR25)}} & 134.1 ($\times$5.5) & 194.0 ($\times$4.2) & 2522.2 ($\times$5.0) \\
    CDPruner{\small\texttt{(NIPS25)}} & 134.1 ($\times$5.5) & 191.5 ($\times$4.3) & 2538.0 ($\times$5.0) \\
    HiPrune{\small\texttt{(AAAI26)}} & 122.7 ($\times$6.1) & 175.9 ($\times$4.7) & 2522.2 ($\times$5.0) \\
    \textbf{EADP}{\small\texttt{(Ours)}} & \textbf{129.5 ($\times$5.7)} & \textbf{194.2 ($\times$4.2)} & \textbf{2538.2 ($\times$5.0)} \\
    \bottomrule
    \end{tabular}
    }
    \label{tab:efficiency-llava-1.6}
  \end{minipage}%
  \hfill   
  \begin{minipage}[t]{0.475\textwidth}
    \centering
    \captionof{table}{Efficiency analysis on Qwen2.5-VL-3B.}

    \resizebox{\linewidth}{!}{%
    \begin{tabular}{l|ccc}
    \toprule
    \textbf{Method} & \textbf{Prefill(ms)} & \textbf{Latency(ms)} & \textbf{FLOPs(G)} \\
    \rowcolor{lightgray!75}\multicolumn{4}{c}{\textit{Upper Bound, All 1008 Tokens} ($\mathbf{100\%}$)} \\
    \rowcolor{lightgray!25} Qwen2.5-VL-3B & 351.6 & 691.6 & 4273.3 \\
    \rowcolor{lightgray!75}\multicolumn{4}{c}{\textit{Retain 512 Tokens} \textcolor{Green}{($\downarrow\mathbf{49.2\%}$)}} \\
    DivPrune{\small\texttt{(CVPR25)}} & 88.4 ($\times$4.0) & 684.9 ($\times$1.0) & 1854.1 ($\times$2.3) \\
    CDPruner{\small\texttt{(NIPS25)}} & 81.4 ($\times$4.3) & 873.3 ($\times$0.8) & 1854.1 ($\times$2.3) \\
    HiPrune{\small\texttt{(AAAI26)}} & 86.1 ($\times$4.1) & 1883.5 ($\times$0.4) & 1854.1 ($\times$2.3) \\
    \textbf{EADP}{\small\texttt{(Ours)}} & \textbf{79.4 ($\times$4.4)} & \textbf{704.2 ($\times$1.0)} & \textbf{1854.1 ($\times$2.3)} \\
    \rowcolor{lightgray!75}\multicolumn{4}{c}{\textit{Retain 256 Tokens} \textcolor{Green}{($\downarrow\mathbf{74.6\%}$)}} \\
    DivPrune{\small\texttt{(CVPR25)}} & 64.4 ($\times$5.5) & 618.3 ($\times$1.1) & 1064.2 ($\times$4.0) \\
    CDPruner{\small\texttt{(NIPS25)}} & 65.4 ($\times$5.4) & 747.6 ($\times$0.9) & 1064.2 ($\times$4.0) \\
    HiPrune{\small\texttt{(AAAI26)}} & 67.3 ($\times$5.2) & 2019.7 ($\times$0.3) & 1064.2 ($\times$4.0) \\
    \textbf{EADP}{\small\texttt{(Ours)}} & \textbf{62.9 ($\times$5.6)} & \textbf{622.3 ($\times$1.1)} & \textbf{1064.2 ($\times$4.0)} \\
    \rowcolor{lightgray!75}\multicolumn{4}{c}{\textit{Retain 128 Tokens} \textcolor{Green}{($\downarrow\mathbf{87.3\%}$)}} \\
    DivPrune{\small\texttt{(CVPR25)}} & 59.0 ($\times$6.0) & 690.2 ($\times$1.0) & 669.2 ($\times$6.4) \\
    CDPruner{\small\texttt{(NIPS25)}} & 61.5 ($\times$5.7) & 1032.7 ($\times$0.7) & 669.2 ($\times$6.4) \\
    HiPrune{\small\texttt{(AAAI26)}} & 68.4 ($\times$5.1) & 2591.7 ($\times$0.3) & 669.2 ($\times$6.4) \\
    \textbf{EADP}{\small\texttt{(Ours)}} & \textbf{59.7 ($\times$5.9)} & \textbf{587.5 ($\times$1.2)} & \textbf{669.2 ($\times$6.4)} \\
    \bottomrule
    \end{tabular}
    }
    \label{tab:efficiency-qwen2.5-vl}
  \end{minipage}

\end{table}
\begin{table*}[!t]
  \centering
  \caption{\textbf{Time breakdown of EADP on Qwen3-VL-8B.} All times are measured in milliseconds with 1024 input visual tokens. Mean and standard deviation are computed per benchmark and then averaged across 10 benchmarks.}
  \label{tab:wall-time-clock}
  \resizebox{\textwidth}{!}{
    \begin{tabular}{c|c|cccccc|c}
    \toprule
    \textbf{Retained} & \textbf{Baseline} & \multicolumn{6}{c|}{\textbf{EADP}} & \textbf{Pruned} \\
    \textbf{Tokens} & \textbf{Prefill} & \textbf{Global Guidance} & \textbf{Dense Guidance} & \textbf{Score Fusion} & \textbf{Smoothing \& Polarization} & \textbf{Facility Location} & \textbf{Total} & \textbf{Prefill} \\
    \midrule
    128 & $320.51 \mathbin{\pm} 8.47$ & $0.37 \mathbin{\pm} 0.04$ & $0.66 \mathbin{\pm} 0.08$ & $0.29 \mathbin{\pm} 0.02$ & $0.49 \mathbin{\pm} 0.04$ & $37.28 \mathbin{\pm} 1.28$ & $39.09 \mathbin{\pm} 1.36$ & $77.60 \mathbin{\pm} 7.07$ \\
    256 & $320.51 \mathbin{\pm} 8.47$ & $0.37 \mathbin{\pm} 0.04$ & $0.65 \mathbin{\pm} 0.08$ & $0.29 \mathbin{\pm} 0.03$ & $0.49 \mathbin{\pm} 0.05$ & $74.26 \mathbin{\pm} 5.84$ & $76.05 \mathbin{\pm} 5.98$ & $122.63 \mathbin{\pm} 3.64$ \\
    512 & $320.51 \mathbin{\pm} 8.47$ & $0.37 \mathbin{\pm} 0.04$ & $0.66 \mathbin{\pm} 0.09$ & $0.28 \mathbin{\pm} 0.03$ & $0.49 \mathbin{\pm} 0.05$ & $143.73 \mathbin{\pm} 9.94$ & $145.53 \mathbin{\pm} 10.06$ & $178.97 \mathbin{\pm} 14.00$ \\
    \bottomrule
    \end{tabular}
  }
\end{table*}

\section{More Ablations and Details}
\label{sec:supp_more_ablation}

\noindent \textbf{Details of aggregation method.} Here, we provide detailed descriptions of the four aggregation variants reported under ``Aggregation method” in Table 6 of the main paper.
\textbf{(1)} Using the weighted inverse entropy. We replace Eq. (7) in the main text with $\alpha_{i,j} = \texttt{softmax}_{i\in\mathcal{T}'}(c_{i,j}/h_i* \gamma)$, where $c_{i,j}$ denotes the cosine similarity between text token $t_i$ and visual token $v_j$, i.e., the similarity values computed in Eq. (3) of our main paper.
\textbf{(2)} Averaging over the top-k scores. Instead of aggregating dense guidance scores from all retained text tokens, we only use the dense guidance scores associated with the top-k retained text tokens.
\textbf{(3)} Using the gated L1 norm. We modify Eq. (7) to $\alpha_{i} = \texttt{softmax}_{i\in\mathcal{T}'}(\theta \cdot (\mu_H - h_i))$, where $\mu_H$ denotes the mean entropy value across all text tokens.

\noindent \textbf{Ablation on the temperature parameters.} We evaluate different temperature values $\tau$ and $\gamma$ to validate our choices of Eq. (4) and Eq. (7) in our main paper, respectively. As shown in~\cref{tab:temp_ablation}, setting the softmax temperature in Eq. (4) to $\tau=0.01$ yields the best performance; when $\tau$ becomes smaller, performance drops noticeably. We therefore use $\tau=0.01$ as the default. For Eq. (7), varying the temperature $\gamma$ leads to only minor differences, indicating that the method is generally robust to this hyperparameter. We thus also set $\gamma=0.01$ by default.

\noindent \textbf{Ablation on the dynamic allocation manner.} As shown in~\cref{tab:dynamic-allocation}, the two variants achieve comparable performance. We therefore adopt the first option, as it is more implementation-efficient and only requires a summation operation to allocate the retention budget, i.e., the number of tokens to keep for each patch.

\begin{table*}[t]
\caption{Ablation studies on temperature coefficients. All experiments are conducted using LLaVA-1.5-7B with a fixed budget of 128 visual tokens}
\label{tab:temp_ablation}
\centering

\begin{minipage}[t]{0.49\textwidth}
    \centering
    \resizebox{\linewidth}{!}{
    \begin{tabular}{lcccccc}
    \toprule
     & \textbf{VizWiz} & \textbf{SQA} & \textbf{TextVQA} & \textbf{POPE} & \textbf{MME} & \textbf{Avg.} \\
    \midrule
    \rowcolor{lightgray!75}\multicolumn{7}{c}{Temperature coefficient 1} \\
    $\tau = 0.1$ & 57.7 & 68.9 & 56.4 & 86.9 & 1425.5 & 68.2 \\
    $\tau = 0.02$ & 57.8 & 69.0 & 56.4 & 87.1 & 1423.1 & 68.3 \\
    $\tau = 0.01$ & 58.0 & 69.0 & 56.5 & 87.2 & 1439.0 & 68.5 \\
    $\tau = 0.005$ & 57.7 & 68.8 & 56.1 & 86.8 & 1421.9 & 68.1 \\
    $\tau = 0.002$ & 57.4 & 68.7 & 55.8 & 86.6 & 1397.4 & 67.6 \\
    \bottomrule
    \end{tabular}
    }
\end{minipage}
\hfill

\begin{minipage}[t]{0.49\textwidth}
    \centering
    \resizebox{\linewidth}{!}{
    \begin{tabular}{lcccccc}
    \toprule
     & \textbf{VizWiz} & \textbf{SQA} & \textbf{TextVQA} & \textbf{POPE} & \textbf{MME} & \textbf{Avg.} \\
    \midrule
    \rowcolor{lightgray!75}\multicolumn{7}{c}{Temperature coefficient 2} \\
    $\gamma = 0.1$ & 57.9 & 69.0 & 56.4 & 87.3 & 1441.1 & 68.5 \\
    $\gamma = 0.02$ & 57.7 & 68.8 & 56.4 & 87.3 & 1440.8 & 68.5 \\
    $\gamma = 0.01$ & 58.0 & 69.0 & 56.5 & 87.2 & 1439.0 & 68.5 \\
    $\gamma = 0.005$ & 57.8 & 68.8 & 56.4 & 87.1 & 1438.7 & 68.4 \\
    $\gamma = 0.002$ & 58.0 & 69.0 & 56.4 & 87.1 & 1445.2 & 68.5 \\
    \bottomrule
    \end{tabular}
    }
\end{minipage}

\end{table*}

\begin{table}[t]
    \centering
    \caption{\textbf{Dynamic allocation ablations on LLaVA-NEXT-7B.} Avg. denotes the average performance across 10 benchmarks. Overall stands for average performance across 3 token budgets.}
    \label{tab:dynamic-allocation}
    \resizebox{\linewidth}{!}{
      \begin{tabular}{l|cccccccccc|c}
      \toprule
      \textbf{Token Budget} & \textbf{$\text{VQA}^\text{V2}$} & \textbf{GQA} & \textbf{VizWiz} & \textbf{$\text{SQA}^\text{IMG}$} & \textbf{$\text{VQA}^\text{Text}$} & \textbf{POPE} & \textbf{MME} & \textbf{$\text{MMB}^\text{EN}$} & \textbf{$\text{MMB}^\text{CN}$} & \textbf{MMVet} & \textbf{Avg.} \\
      \rowcolor{lightgray!75}\multicolumn{12}{c}{\textit{Importance-Based Allocation}} \\
      \textbf{160 tokens} & 77.0 & 61.6 & {60.2}$^*$ & 67.4 & 54.2 & 86.0 & 1419.5 & 63.4 & 54.5 & 35.6 & 63.1 \\
      \textbf{320 tokens} & 78.6 & 62.2 & {60.4}$^*$ & 67.6 & 57.0 & 86.9 & 1491.3 & 64.6 & 55.9 & 39.4 & 64.7 \\
      \textbf{640 tokens} & 80.0 & 62.7 & {60.5}$^*$ & 68.0 & 59.2 & 87.4 & 1494.7 & 66.5 & 57.9 & 40.1 & 65.7 \\
      \rowcolor{lightgray!25}\textbf{Overall} & 78.5 & 62.2 & 60.4 & 67.7 & 56.8 & 86.8 & 73.4 & 64.8 & 56.1 & 38.4 & 64.5 \\
      \rowcolor{lightgray!75}\multicolumn{12}{c}{\textit{Global-Guided Allocation}} \\
      \textbf{160 tokens} & 77.0 & 61.6 & {59.6}$^*$ & 67.7 & 54.2 & 86.4 & 1445.2 & 62.6 & 54.2 & 34.7 & 63.0 \\
      \textbf{320 tokens} & 78.6 & 62.3 & {60.4}$^*$ & 67.1 & 56.8 & 83.7 & 1501.7 & 64.9 & 57.0 & 38.5 & 64.4 \\
      \textbf{640 tokens} & 80.0 & 62.8 & {60.6}$^*$ & 68.3 & 58.9 & 86.3 & 1504.0 & 66.5 & 57.6 & 41.5 & 65.8 \\
      \rowcolor{lightgray!25}\textbf{Overall} & 78.5 & 62.2 & 60.2 & 67.7 & 56.6 & 85.5 & 74.2 & 64.7 & 56.2 & 38.2 & 64.4 \\
      \bottomrule
      \end{tabular}
    }
\end{table}


%
%
\bibliographystyle{splncs04}
\bibliography{main}
\end{document}